%% file: main.tex
\documentclass{article}
\usepackage{fancyhdr}

    \usepackage{neurips_2024}

\usepackage{soul}
\usepackage[utf8]{inputenc} 
\usepackage[T1]{fontenc}    
\usepackage{hyperref}       
\usepackage{url}            
\usepackage{booktabs}       
\usepackage{amsfonts}       
\usepackage{fancyhdr}       

\usepackage{nicefrac}       
\usepackage{microtype}      
\usepackage{graphicx}
\usepackage{pifont}
\usepackage{multirow}
\usepackage{CJKutf8}
\definecolor{darkmagenta}{rgb}{0.56, 0.0, 1.0}
\definecolor{softyellow}{rgb}{1.0, 0.92, 0.3} 
\definecolor{LightAquamarine}{rgb}{0.75, 1.0, 0.8} 
\definecolor{FireBrick}{RGB}{178,34,34}
\definecolor{MediumPurple}{RGB}{147,112,219}

\definecolor{uclablue}{rgb}{0.15, 0.45, 0.68}
\hypersetup{
    breaklinks,
    colorlinks=true,
    citecolor={darkmagenta},
    linkcolor={uclablue},
    urlcolor={uclablue}
}
\usepackage{wrapfig}
\usepackage{float}
\usepackage{subcaption}
\usepackage{placeins}
\usepackage{lipsum} 
\usepackage{tcolorbox}
\usepackage{amsmath}
\usepackage{amssymb}
\usepackage{utfsym}
\usepackage{fontawesome}
\usepackage{xspace}
\usepackage{enumitem}
\usepackage{multirow} 
\tcbuselibrary{breakable}
\usepackage{enumitem}
\usepackage{colortbl}
\usepackage{fancyhdr}

\usepackage{tcolorbox}
\usepackage{transparent}

\newtcolorbox{abstractbox}{
    colback=blue!5!white,     
    frame empty,   
    boxrule=1pt,              
    arc=4mm,                  
    left=8pt,                 
    right=8pt,                
    top=8pt,                  
    bottom=8pt,                
    opacityback=0.9
}

\title{Part \MakeUppercase{\romannumeral 1}: Tricks or Traps? \\[5pt]
A Deep Dive into RL for LLM Reasoning}

    \vspace{-0.5cm}

\author{
\textbf{Zihe Liu}$^{*}$\textsuperscript{$\heartsuit$ \scalebox{1.3}{$\alpha$}},
\textbf{Jiashun Liu}$^{*}$\textsuperscript{\scalebox{1.3}{$\diamond$}\scalebox{1.3}{$\alpha$}},
\textbf{Yancheng He}$^{*}$\textsuperscript{\scalebox{1.3}{$\alpha$}},
\textbf{Weixun Wang}$^{*\dag}$\textsuperscript{\scalebox{1.3}{$\alpha$}},
\textbf{Jiaheng Liu}\textsuperscript{$\Omega$},
\\[3pt]
\textbf{Ling Pan}\textsuperscript{\scalebox{1.3}{$\diamond$}},
\textbf{Xinyu Hu}\textsuperscript{\scalebox{1.3}{$\alpha$}$\P$}, 
\textbf{Shaopan Xiong}\textsuperscript{\scalebox{1.3}{$\alpha$}}, 
\textbf{Ju Huang}\textsuperscript{\scalebox{1.3}{$\alpha$}}, 
\textbf{Jian Hu}\textsuperscript{$\clubsuit$}, 
\textbf{Shengyi Huang}\textsuperscript{\ddag}, 
\\[3pt]
\textbf{Johan Obando-Ceron}\textsuperscript{\scalebox{1.3}{$\Psi$}}, \textbf{Siran Yang}\textsuperscript{\scalebox{1.3}{$\alpha$}}, 
\textbf{Jiamang Wang}\textsuperscript{\scalebox{1.3}{$\alpha$}}, 
\textbf{Wenbo Su}\textsuperscript{\scalebox{1.3}{$\alpha$}}, 
\textbf{Bo Zheng}\textsuperscript{\scalebox{1.3}{$\alpha$}} \\[6pt]
\textsuperscript{\scalebox{1.3}{$\alpha$}}Alibaba Group \quad \textsuperscript{$\heartsuit$} Beijing Jiaotong University \\ \textsuperscript{\scalebox{1.3}{$\diamond$}} Hong Kong University of Science and Technology \quad \textsuperscript{$\Omega$} Nanjing University \\ \textsuperscript{$\P$} Peking University \quad \textsuperscript{$\clubsuit$} OpenRLHF \quad \textsuperscript{\ddag} CleanRL \quad \textsuperscript{$\Psi$} Mila\\[6pt]
}

\begin{document}

\maketitle
\let\oldthefootnote\thefootnote

\let\thefootnote\relax\footnotetext{*~Equal Contribution. ~~$^\dagger$~Corresponding to: Weixun Wang <weixun.wwx@taobao.com>.}
\let\thefootnote\oldthefootnote

\input{content/0_Abstract}



\input{content/1_Introduction}
\input{content/2_RelatedWork}

\input{content/3_Preliminaries}
\input{content/4_Baselines}

\input{content/5_Analysis}

\input{content/10_Final_Result}
\input{content/Conclusion}
\bibliographystyle{unsrtnat}
\bibliography{references} 

\input{content/11_Appendix}

\label{sec:appendix}
\end{document}

%% file: content/0_Abstract.tex
\begin{abstractbox}
\begin{center}
\textbf{\Large Abstract}
\end{center}
Reinforcement learning (RL) for LLM reasoning has rapidly emerged as a prominent research area, marked by a significant surge in related studies on both algorithmic innovations and practical applications. Despite this progress, several critical challenges remain, including the absence of standardized guidelines for applying RL techniques and a fragmented understanding of their underlying mechanisms. In addition, inconsistent experimental settings, variations in training data, and differences in model initialization have led to conflicting conclusions, obscuring the key characteristics of these techniques and creating confusion among practitioners when selecting appropriate techniques. This paper systematically reviews widely adopted RL techniques through rigorous reproductions and isolated evaluations within a unified open-source framework. We analyze the internal mechanisms, applicable scenarios, and core principles of each technique through fine-grained experiments, including datasets of varying difficulty, model sizes, and architectures. Based on these insights, we present clear guidelines for selecting RL techniques tailored to specific setups and provide a reliable roadmap for practitioners navigating the RL for the LLM domain. Finally, we show that a minimalist combination of two techniques can unlock the learning capability of critic-free policies with a vanilla PPO loss. The results demonstrate that our simple combination consistently improves performance, surpassing strategies such as GRPO and DAPO.
\end{abstractbox}

\begin{figure}[!h]
    \centering
    \includegraphics[width=0.9\linewidth]{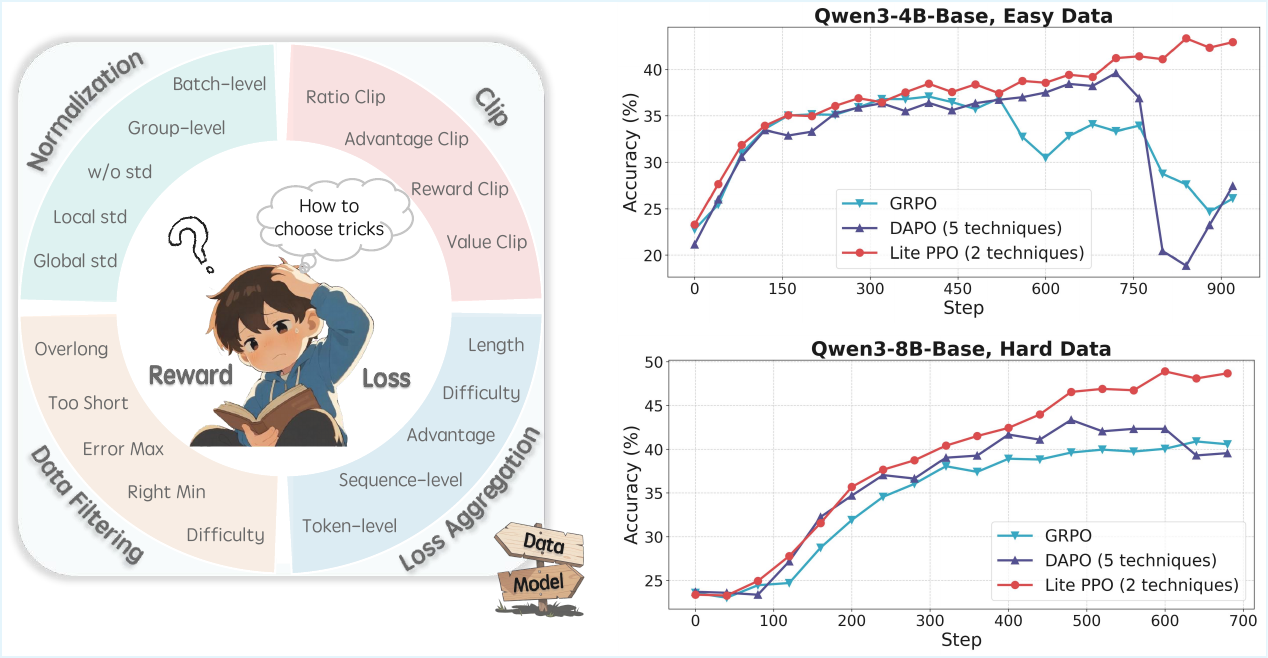}
    \caption{\textbf{Left}: The proliferation of RL optimization techniques, coupled with diverse initialized models and data, has raised barriers to practical adoption. \textbf{Right}: We establish detailed application guidelines via dissecting internal mechanisms of widely-used tricks, and introduce \textcolor{FireBrick}{Lite PPO}, a minimalist two-technique combination that enhances learning capacity in critic-free policies with vanilla PPO loss. The average accuracy is calculated across six mathematical benchmarks.}
    \label{fig:intro}
\end{figure}

%% file: content/1_Introduction.tex
\section{Introduction}
Recent breakthroughs in large language models (LLMs) such as OpenAI o1 \citep{Wu2024ACS} and DeepSeek R1 \citep{shao2024deepseekmath} have positioned reinforcement learning (RL) as a key driver in unlocking advanced reasoning capabilities in LLMs. This is particularly evident in challenging reasoning tasks like mathematical problem solving \citep{he2025deepmath} and code generation \citep{zhuo2024bigcodebench}, where RL has demonstrated the potential to elevate LLM performance beyond what pre-training alone can achieve. Such an emerging trend has sparked widespread interest within the research community in the direction of "RL for LLM" (or RL4LLM). In 2025, RL4LLM experienced explosive growth, producing hundreds of publications across arXiv and major conferences, covering a wide range of topics from algorithmic innovation to practical engineering solutions.

However, this rapid progress is shadowed by the lack of usage guidelines for existing RL techniques and tricks \citep{Huang2024TheNI} as well as the absence of in-depth analysis of their underlying mechanisms. Specifically, these limitations can manifest as confusion among practitioners in choosing RL tricks, as different papers advocate conflicting solutions to the same problem. For instance, GRPO \citep{shao2024deepseekmath} advocates for group-level normalization to enhance policy stability, whereas REINFORCE++ \citep{hu2025reinforce++} argues that batch-level normalization works better. Moreover, GRPO incorporates variance in normalization, yet Dr. GRPO \citep{liu2025understanding} explicitly recommends removing variance normalization to prevent bias. Similarly, GRPO \citep{shao2024deepseekmath} has achieved a breakthrough in performance through the strategy of using response-level loss calculation, while DAPO \citep{yu2025dapo} has instead adopted token-level loss calculation. Such contradictory and chaotic phenomena underscore the fragmented understanding and inconsistent recommendations within the RL4LLM community. A likely cause for the above phenomenon is that the experimental settings, training data, and initialization of the existing work all have significant differences, which may also cause deviations in the summary of the conclusions.

Apart from the confusion caused by the intrinsic differences of similar techniques, the numerous and seemingly orthogonal techniques, including \textit{Normalization, Clip, and Overlong Filtering}, also increase the complexity of algorithm application in practice. Practitioners face non-trivial challenges in identifying an effective combination from a wide range of techniques to unlock the learning capacity of LLMs in specific scenarios. These ambiguities have naturally triggered a key requirement of practitioners: 

\textbf{\textcolor{purple}{What scenarios are the existing techniques respectively suitable for? Is there a simple and generalizable combination that can be used to enhance policy optimization?}}

Aligned with classic RL mechanism analysis methodologies \citep{Andrychowicz2020WhatMI, Engstrom2020ImplementationMI, Huang2024TheNI}, we systematically review the widely used RL techniques by reproducing them and independently evaluating the actual impact of each technique, based on the same open-source infrastructure framework and policy models. To comprehensively cover practical scenarios, we design extensive experimental settings incorporating datasets of varying difficulty levels, diverse model sizes, and distinct model types. Furthermore, we conduct an in-depth analysis of their theoretical foundations, implementation details, and applicable scenarios as demons. The intuitive contribution is illustrated in Figure~\ref{fig:intro}. Specifically, \ding{182} our empirical results reveal that most RL techniques exhibit strong preferences and sensitivities to the experimental setup, e.g., model type, data distribution, reward mechanism and hyperparameter. \ding{183} Based on the isolated analysis under our setup, we demonstrate that employing just two techniques, i.e., \textit{advantage normalization (group-level mean, batch-level std) and token-level loss aggregation}, can unlock the learning capability of critic-free policies using vanilla PPO loss, surpassing mainstream RL4LLM algorithms incorporating redundant components. Our core contributions are selected as:
\definecolor{blueviolet}{RGB}{138,43,226}
\newtcolorbox{insightblock}{
  colback=blueviolet!5,   
  colframe=blueviolet!50!black!50!,    
  boxrule=0.5mm,       
  arc=2mm,             
  left=0pt,            
  right=8pt,           
  top=8pt,             
  bottom=8pt,          
}

\begin{insightblock}
\begin{enumerate}[leftmargin=1.5em]
    \item Group-level normalization shows robust efficiency under each reward setting. Batch-level normalization provides more stable improvement under large scale reward setting. (\S\ref{sec.5.1.1})
    \item Group-level mean and batch-level standard deviation enable further robust normalization. (\S\ref{sec.5.1.3})
    \item Clip Higher promotes high-quality exploration for aligned models. (\S\ref{sec.5.2.1})
    \item There appears to be a ``scaling law'' between the performance and the upper bound of the clipping on the small-sized model. (\S\ref{sec.5.2.3})
    \item Compared to sequence-level loss aggregation, token-level aggregation is effective on base models but shows limited improvement on aligned models. (\S\ref{sec.5.3.1})
    \item Overlong filtering enhances accuracy and clarity for short-to-medium reasoning tasks but provides limited benefits for long-tail reasoning. (\S\ref{sec.5.4.1})
    \item Two techniques may unlock learning capacity in critic-free policies based on vanilla PPO loss. (\S\ref{sec.6.1.1})
\end{enumerate}
\end{insightblock}


%% file: content/2_RelatedWork.tex

%% file: content/3_Preliminaries.tex
\section{Preliminaries}

\subsection{Proximal Policy Optimization (PPO)}

Proximal Policy Optimization (PPO)\citep{schulman2017proximal} is a widely used actor-critic algorithm grounded in the policy gradient framework. It improves the stability of policy learning by optimizing a clipped surrogate objective that restricts the divergence between the new and old policies during training. The PPO objective is: 

\begin{equation}
\begin{aligned}
\mathcal{J}_{\mathrm{PPO}}(\theta) =\ 
&\mathbb{E}_{\left[ 
  q \sim P(Q),\ 
  o \sim \pi_{\theta_{\mathrm{old}}}(O|q)
\right]} \\
&
  \frac{1}{|o|} \sum_{t=1}^{|o|}
  \min\Bigg(
    \frac{\pi_\theta(o_t|q, o_{<t})}{\pi_{\theta_{\mathrm{old}}}(o_t|q, o_{<t})} A_t,\, 
    \mathrm{clip}\left(
      \frac{\pi_\theta(o_t|q, o_{<t})}{\pi_{\theta_{\mathrm{old}}}(o_t|q, o_{<t})},\, 1{-}\epsilon,\, 1{+}\epsilon
    \right) A_t
  \Bigg),
\end{aligned}
\end{equation}

where $\pi_\theta$ and $\pi_{\theta_\text{old}}$ denote the current and old policy models, respectively. $q$ and $o$ represent the sampled question and output sequence, with $o_t$ as the $t$-th token in $o$. $\epsilon$ is a clipping hyperparameter for stabilizing updates. $A_t$ is the advantage at step $t$, typically estimated via Generalized Advantage Estimation (GAE)~\citep{schulman2018highdimensionalcontinuouscontrolusing}. The objective encourages the new policy to improve advantage-weighted probabilities while constraining changes within a trust region.

\subsection{Group Relative Policy Optimization (GRPO)}
Group Relative Policy Optimization (GRPO), proposed in DeepSeekMath~\citep{shao2024deepseekmath}, eliminates the value function (critic) and instead estimates the advantage by normalizing rewards within a group of sampled responses for the same prompt. Specifically, for a prompt $x$ with $G$ responses and associated rewards $\{r_i\}_{i=1}^G$, the group-normalized advantage is given by:
\begin{equation}
\hat{A}_{i,t} = \frac{r_i - \mathrm{mean}(\{r_i\}_{i=1}^G)}{\mathrm{std}(\{r_i\}_{i=1}^G)}.
\end{equation}
The effectiveness of the above normalization method can be understood from the perspective of reward shaping. By emphasizing the differences among candidate outputs for the same prompt, it effectively preserves the reliability of the gradient signal, even in sparse reward settings~\citep{Hu2020LearningTU}. Instead of adding a KL penalty to the reward, GRPO directly regularizes by directly adding the KL divergence between the trained policy and the reference policy to the loss. The overall surrogate objective is:

\begin{equation}
\begin{aligned}
\mathcal{J}_{\mathrm{GRPO}}(\theta) =\ 
&\mathbb{E}_{\left[ 
  q \sim P(Q),\, \{o_i\}_{i=1}^G \sim \pi_{\theta_{\mathrm{old}}}(O|q)
\right]} \\
&
  \frac{1}{G} \sum_{i=1}^G\, \frac{1}{|o_i|} \sum_{t=1}^{|o_i|}\ 
  \left\{\min\left(
    r_{i,t}(\theta)\, \hat{A}_{i,t},\, 
    \mathrm{clip}\left(
      r_{i,t}(\theta),\, 1{-}\epsilon,\, 1{+}\epsilon
    \right)\, \hat{A}_{i,t}
  \right)
  - \beta D_{\mathrm{KL}}\left[\pi_\theta\, \|\, \pi_{\mathrm{ref}}\right]
\right\},
\end{aligned}
\end{equation}

where $r_{i,t}(\theta) = \frac{\pi_\theta(o_{i,t}|q, o_{i,<t})}{\pi_{\theta_\mathrm{old}}(o_{i,t}|q, o_{i,<t})}$, $\epsilon$ and $\beta$ are hyper-parameters, and $D_\mathrm{KL}$ denotes the KL divergence between the learned policy and a reference policy $\pi_{\mathrm{ref}}$. 

\subsection{Decoupled Clip and Dynamic Sampling Policy Optimization (DAPO)}

Decoupled Clip and Dynamic Sampling Policy Optimization (DAPO)~\citep{yu2025dapo} is a recent RL method designed to address the unique challenges in LLM reasoning. For each question $q$ with gold answer $a$, DAPO samples a group of $G$ outputs $\{o_i\}_{i=1}^G$ from the old policy, computes their rewards, and maximizes the following surrogate objective:

\begin{equation}
\begin{aligned}
\mathcal{J}_{\mathrm{DAPO}}(\theta) =\ 
&\mathbb{E}_{\left[ 
  (q, a) \sim \mathcal{D},\, \{o_i\}_{i=1}^G \sim \pi_{\theta_{\mathrm{old}}}(\cdot|q)
\right]} \\
&
  \frac{1}{\sum_{i=1}^G |o_i|} \sum_{i=1}^G \sum_{t=1}^{|o_i|}\
  \left\{
    \min\left(
      r_{i,t}(\theta)\, \hat{A}_{i,t},\, 
      \mathrm{clip}\left(
        r_{i,t}(\theta),\, 1{-}\epsilon_{\mathrm{low}},\, 1{+}\epsilon_{\mathrm{high}}
      \right)\, \hat{A}_{i,t}
    \right)
  \right\},
\end{aligned}
\end{equation}
where $\hat{A}_{i,t}$ is the group-normalized advantage.  
In addition, DAPO decouples the upper and lower clipping ranges ($\epsilon_\mathrm{low}$, $\epsilon_\mathrm{high}$) to better support exploration, dynamically filters out samples where all responses are correct or incorrect, aggregates losses at the token level, and applies special reward shaping for overlong or truncated responses.

\subsection{Reinforcement Learning Techniques}

A variety of practical techniques have been introduced to stabilize optimization, reduce variance, and accelerate the convergence of LLMs on reasoning tasks. Drawing from prior research and practical implementations, we categorize commonly used techniques as follows.

\paragraph{Baseline Design.}
Baselines are crucial for reducing variance in policy gradient estimation.  
Recent studies have proposed more effective formulations, such as using the mean reward within each group as the baseline \citep{shao2024deepseekmath} and computing the baseline for each sample as the average gradient estimate from other samples in the group \citep{ahmadian2024back,kool2019buy}.

\paragraph{Clipping Strategies.}
Clipping controls excessive updates in policy optimization and can be applied to rewards, advantages, or ratios. Furthermore, the \emph{Clip Ratio Higher} \citep{yu2025dapo} method relaxes the upper bound in PPO’s ratio clipping to better preserve exploration.

\paragraph{Normalization Strategies.}
Normalization of rewards or advantages helps stabilize gradient magnitudes.  
Representative approaches include:  
\emph{Batch-level Reward Normalization} \citep{hu2025reinforce++},  
\emph{Group-level Reward Normalization} \citep{shao2024deepseekmath,ahmadian2024back}, and  
\emph{Reward Shift without Standard Deviation} \citep{liu2025understanding}, which omits the standard deviation term to avoid difficulty bias.

\paragraph{Filtering Strategies.}
Filtering excludes uninformative or undesirable samples prior to gradient computation.  
Examples include:  
\emph{Overlong Filtering} \citep{yu2025dapo} to remove responses exceeding predefined length limits;  
\emph{Error Max Clip Mask} and \emph{Right Min Clip Mask} to filter overly incorrect or trivially correct samples;  
and \emph{Difficulty Mask} \citep{yu2025dapo,Zhang2025SRPOAC,Chu2025GPGAS} to exclude samples outside a targeted difficulty range.

\paragraph{Loss Aggregation Granularity.}
The formulation of loss aggregation determines the relative contribution of each token to the overall objective. Common approaches include:
\emph{Token-level Loss} computes per-token advantages to reduce length bias, while
\emph{Sequence-level Loss} aggregates at the sequence level.

\paragraph{Additional Loss Functions.}
Auxiliary losses can complement the primary objective and regularize training.  
\emph{KL Loss} \citep{yu2025dapo,liu2025understanding} constrains divergence from a reference policy,  
while \emph{SFT Loss} \citep{Zhang2025GRPOLEADAD} incorporates supervised fine-tuning objectives to preserve alignment.

\paragraph{Reward Design.}
Shaping the reward function can guide desired output properties. Common examples include:
\emph{Length Penalty} discourages excessively long outputs;
\emph{Formatting Reward} which encourages outputs that adhere to preferred structures such as boxed answers, bullet lists, or code-style formatting;
\emph{Length-Dependent Accuracy Reward} combines correctness with output length.

\medskip
\noindent
These categories summarize the most prevalent strategies for improving RL in LLM reasoning.  
In this work, we focus on four key aspects: \emph{Normalization}, \emph{Clipping}, \emph{Masking}, and \emph{Loss Aggregation}, and conduct in-depth analyses of their mechanisms and practical utility.

%% file: content/4_Baselines.tex
\section{Experimental Designs}
\subsection{Experimental Setup}
\paragraph{Training Algorithm:}
We utilize the open-sourced ROLL framework\footnote{Open source RL framework: \url{https://github.com/alibaba/ROLL}}~\citep{Wang2025ReinforcementLO}, an efficient and scalable platform specifically designed for reinforcement learning optimization in LLMs, to conduct all experiments. In addition, we adopt the PPO loss \citep{schulman2017proximal}, with advantage values computed using the REINFORCE algorithm \citep{NIPS1999_464d828b} as the unified RL baseline. To ensure consistency with prior research, we set the global batch size to $1024$ by using a rollout batch size of $128$ and sampling $8$ responses per prompt, with a maximum response length of $8192$ tokens.
The learning rate is set to $1e-6$. For text generation, we use a top\_p value of $0.99$, a top\_k value of $100$, and a temperature of $0.99$.

\paragraph{Base Models:}
To comprehensively evaluate reinforcement learning (RL) techniques across parameter scales, our experiments cover two model sizes: Qwen3-4B and Qwen3-8B. For each model size, we include both non-aligned pre-trained versions (Qwen3-4B-Base and Qwen3-8B-Base) and aligned versions, enabling assessment RL gains from different initialization conditions\footnote{Checkpoint links: \url{https://huggingface.co/Qwen/Qwen3-4B}; \url{https://huggingface.co/Qwen/Qwen3-8B}; \url{https://huggingface.co/Qwen/Qwen3-4B-Base; https://huggingface.co/Qwen/Qwen3-8B-Base}}.
    
\begin{wrapfigure}{r}{0.45\textwidth} 
  \vspace{-1cm}  
  \centering
  \includegraphics[width=\linewidth]{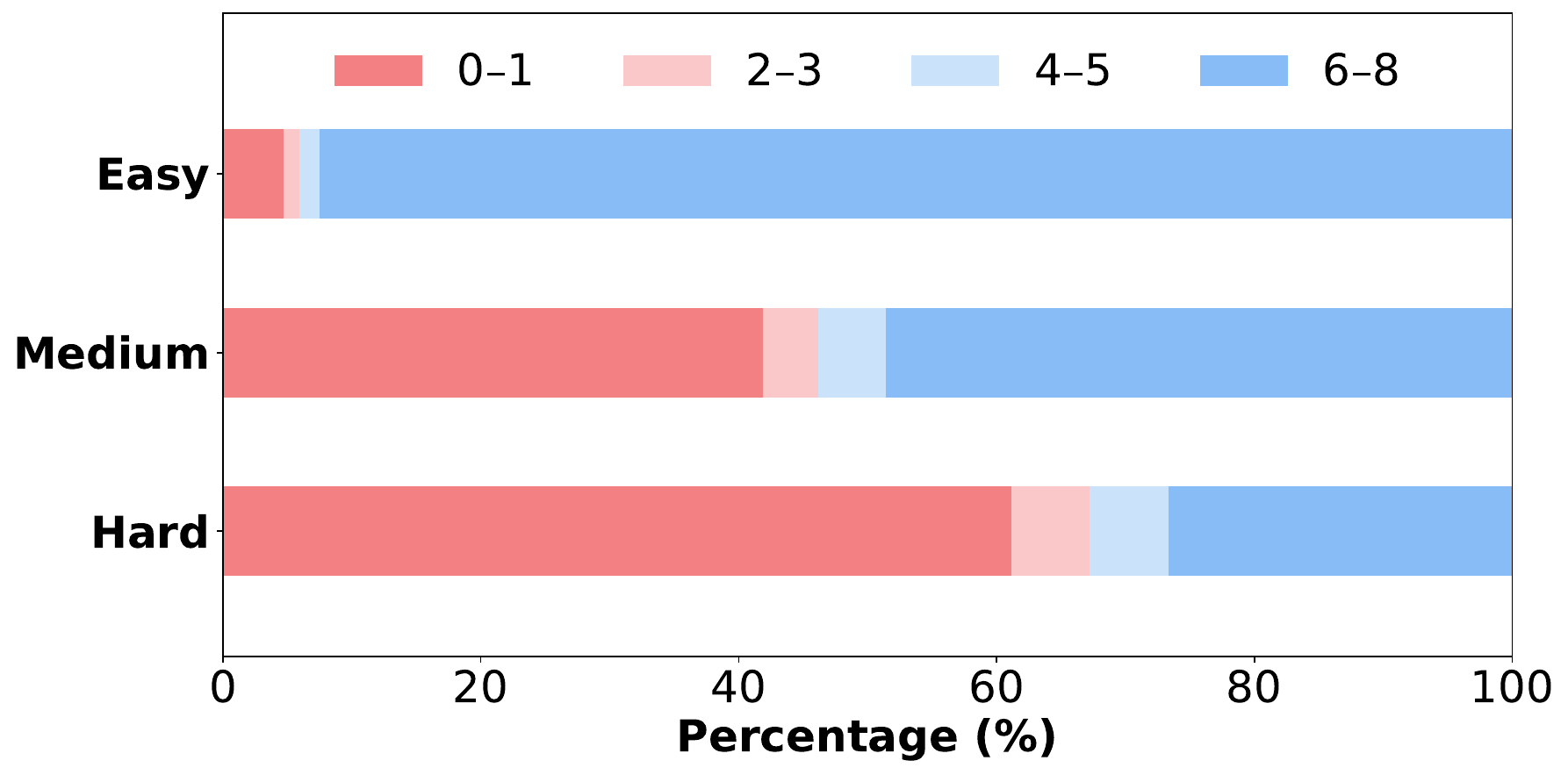}
  \caption{Number of correct responses under 8 rollout iterations across datasets.}
  \label{fig:difficulty}
  \vspace{-0.5cm}
\end{wrapfigure}

\paragraph{Training Datasets:}
To ensure reproducibility and fairness, we exclusively use open-source datasets for training, including \textit{SimpleRL-Zoo-Data} \citep{zeng2025simplerl} and \textit{Deepmath} \citep{he2025deepmath}. To comprehensively examine how problem difficulty affects the RL technique's performance, we randomly sample from the datasets, removing  an excessive proportion of examples whose ground-truth label is simply “True” or “False”. 
This adjustment addresses the \textbf{ostensible positive phenomenon}, where models produce correct binary answers from erroneous reasoning chains, thereby introducing noisy supervision that compromises training quality (please refer to Appendix~\ref{app_false_postive} for case studies). Figure~\ref{fig:difficulty} visualizes the difficulty across the training dataset assessed by GPT-4o \citep{Hurst2024GPT4oSC}.

\begin{itemize}
\item Easy Data : We randomly sample $5,000$ entries from SimpleRL-Zoo-Data-Easy, which consists of problems drawn from GSM8K and MATH-500-level1.
\item Medium Data: We select the $5,000$ easiest examples from the \textit{DeepMath-103k} dataset, based on their assigned difficulty annotations.
\item Hard Data: We randomly sample $5,000$ entries from \textit{DeepMath-103k}, with sampling probability proportional to each entry’s assigned difficulty level.
\end{itemize}

\paragraph{Evaluation Benchmark:}
All the experiments are conducted on six math datasets: MATH-500~\citep{DBLP:conf/iclr/HendrycksBBZMSS21}, OlympiadBench~\citep{DBLP:conf/acl/HeLBHTSHHHZLQL024}, MinervaMath~\citep{DBLP:conf/nips/LewkowyczADDMRS22}, and subsets of standardized examinations (AIME24-25, AMC23). These datasets span a broad complexity spectrum from basic arithmetic to competition-level mathematics, enabling a comprehensive evaluation of reasoning capabilities.

\subsection{Baseline Results}

\begin{figure}[!h] 
\centering
\textbf{Overview of training accuracy and response length}
\includegraphics[width=1\linewidth]{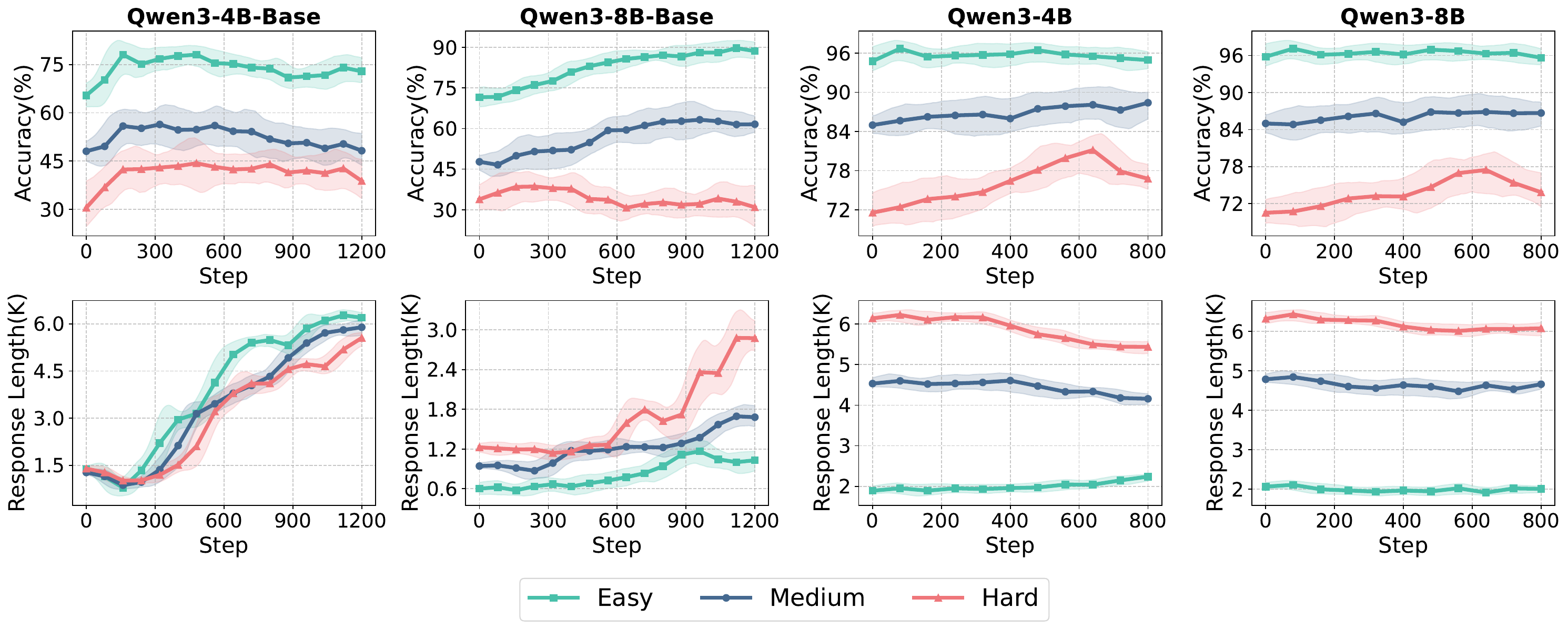}
\textbf{Test accuracy of Base models}
\includegraphics[width=1\linewidth]{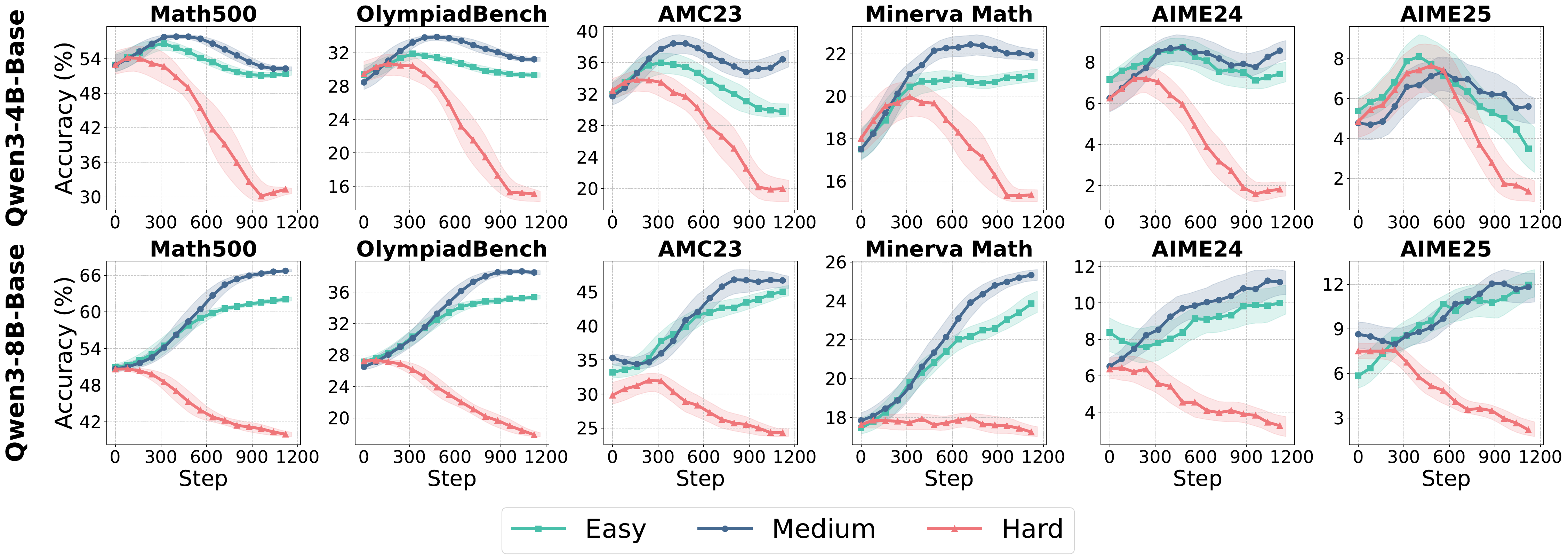}
\textbf{Test accuracy of Aligned models}
\includegraphics[width=1\linewidth]{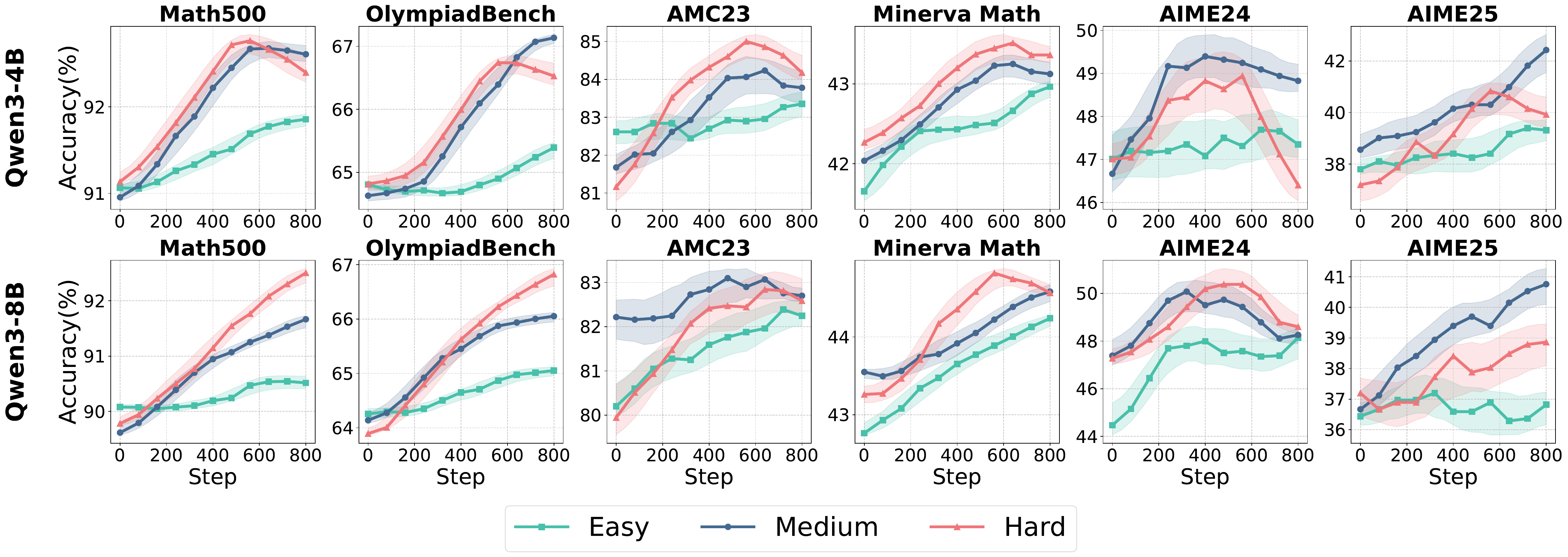}
\caption{\textbf{(Top 2 rows): Test accuracy and response length of four model variants:} \texttt{Qwen3-4B-Base}, \texttt{Qwen3-8B-Base}, \texttt{Qwen3-4B}, and \texttt{Qwen3-8B} across different data difficulty. \textbf{Middle 2 rows}: Accuracy over training iterations of Base models. The first row presents results of \texttt{Qwen3-4B-Base}. The second row shows results of \texttt{Qwen3-8B-Base}. \textbf{Bottom 2 rows}: Accuracy over training iterations of aligned models. The first row presents results of \texttt{Qwen3-4B}, while the second row shows results of \texttt{Qwen3-8B}. To ensure clarity and intuitiveness in the qualitative analysis, all curves are consistently smoothed using identical parameters. Specifically, the mean values are computed using an 11-step moving window with an exponential smoothing factor of \(0.8\). The shaded regions around the curves represent the range \( \text{mean} \pm (\text{std\_multiplier} \times \text{standard deviation}) \), providing a visual representation of the oscillation amplitude.}
\label{baseline_val}
\end{figure}

\paragraph{Impact of Data Difficulty on Training Dynamics} 
We investigate how data difficulty influences the training dynamics of Qwen3 models. 
Specifically, we analyze the training convergence patterns through loss dynamics, accuracy trajectories, and generalization gaps, across three tiers of complexity (\textit{Easy, Medium, Hard}). The detailed learning curves are shown in Figure \ref{baseline_val}.

The experimental results demonstrate that, as the number of training epochs increases, the model exhibits markedly different accuracy trajectories across training sets of different difficulty levels. 
Furthermore, when confronted with more challenging samples, the model often fits complex reasoning patterns by generating more tokens. 


When focusing on the differences in learning efficiency between the unaligned Base model and the aligned model under the same experimental setting (as shown in Figure~\ref{baseline_val}), the aligned models exhibited substantially higher initial accuracy and produced responses with longer average token lengths during early training. However, additional learning yielded only modest gains, with accuracy improving by roughly $2\%$.
This suggests that the current RL4LLM algorithm offers a slight improvement for aligned models that are already highly optimized.



%% file: content/5_Analysis.tex

\section{Analysis}
\subsection{Normalization}
Advantage normalization is a well-established technique for reducing gradient variance and stabilizing policy optimization \citep{zheng2023delve}, and it has become a standard component of RL training pipelines for language models. Yet, substantial differences remain in how it is implemented. For example, GRPO \citep{shao2024deepseekmath} and RLOO \citep{ahmadian2024back,kool2019buy} use group-level normalization, calculating advantages relative to other responses within the same prompt to foster intra-context competition. On the other hand, REINFORCE++ \citep{hu2025reinforce++} employs batch-level normalization, arguing that optimizing within a single prompt excessively can lead to reward hacking and hinder generalization, especially when response diversity is low.

Formally, given a prompt \(x\) with \(K\) sampled responses and corresponding rewards \(\{r_k\}_{k=1}^K\), the group-level normalized advantage for the \(k\)-th response is:
\begin{equation}
A_k^{\mathrm{group}} = \frac{r_k - \mathrm{mean}(\{ r_j \}_{j=1}^{\textcolor{red}{K}})}{\mathrm{std}(\{ r_j \}_{j=1}^{\textcolor{red}{K}})}.
\end{equation}
In contrast, batch normalization computes the reward over a rollout batch of size \(N\) with \(K\) sampled trajectories. The normalized advantage for the \(i\)-th response is:
\begin{equation}
A_i^{\mathrm{batch}} = \frac{r_i - \mathrm{mean}(\{ r_j \}_{j=1}^{\textcolor{blue}{N*K}})}{\mathrm{std}(\{ r_j \}_{j=1}^{\textcolor{blue}{N*K}})}
\end{equation}
\begin{figure}[h] 
\centering
\textbf{Accuracy of 4B-Base model}
\includegraphics[width=1\linewidth]{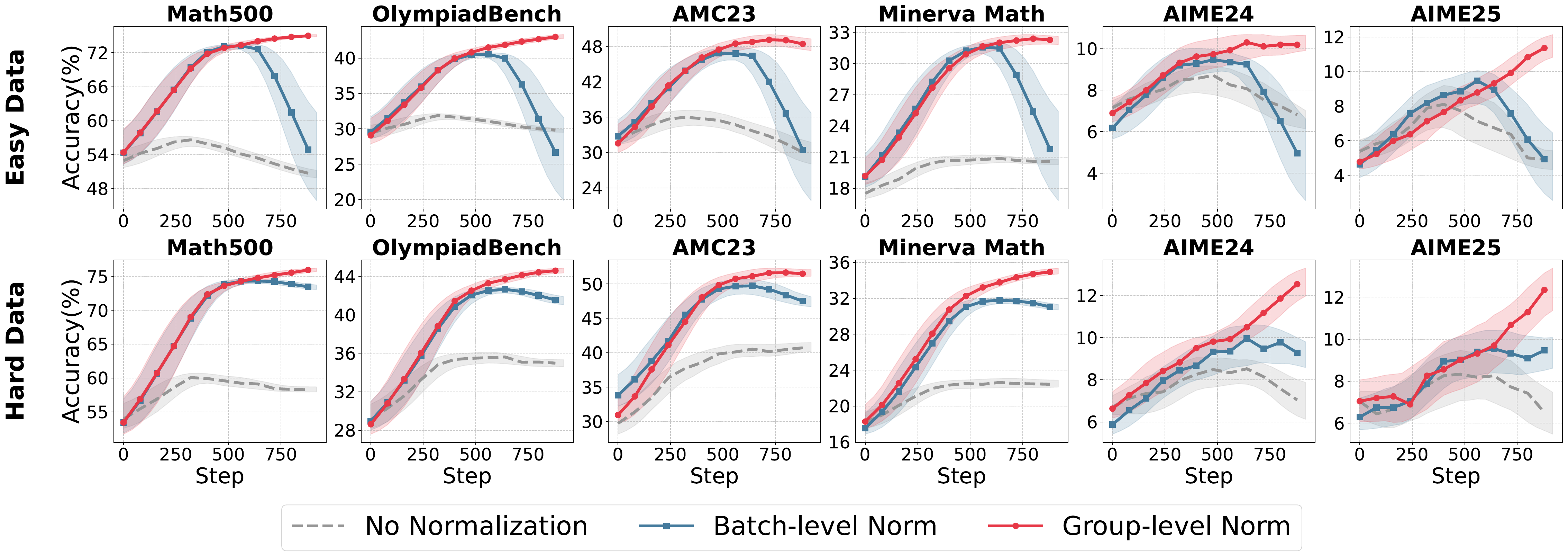}
\textbf{Accuracy of 8B-Base model}
\includegraphics[width=1\linewidth]{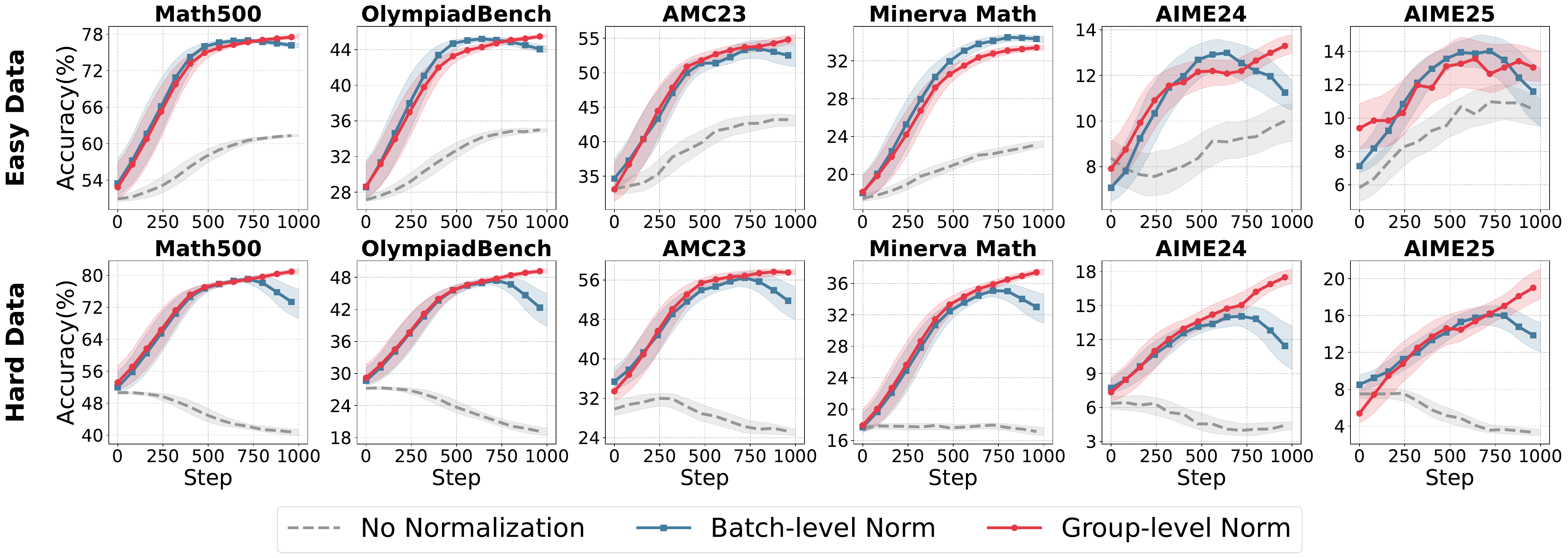}
\textbf{Accuracy of Aligned models}
 \includegraphics[width=1\linewidth]{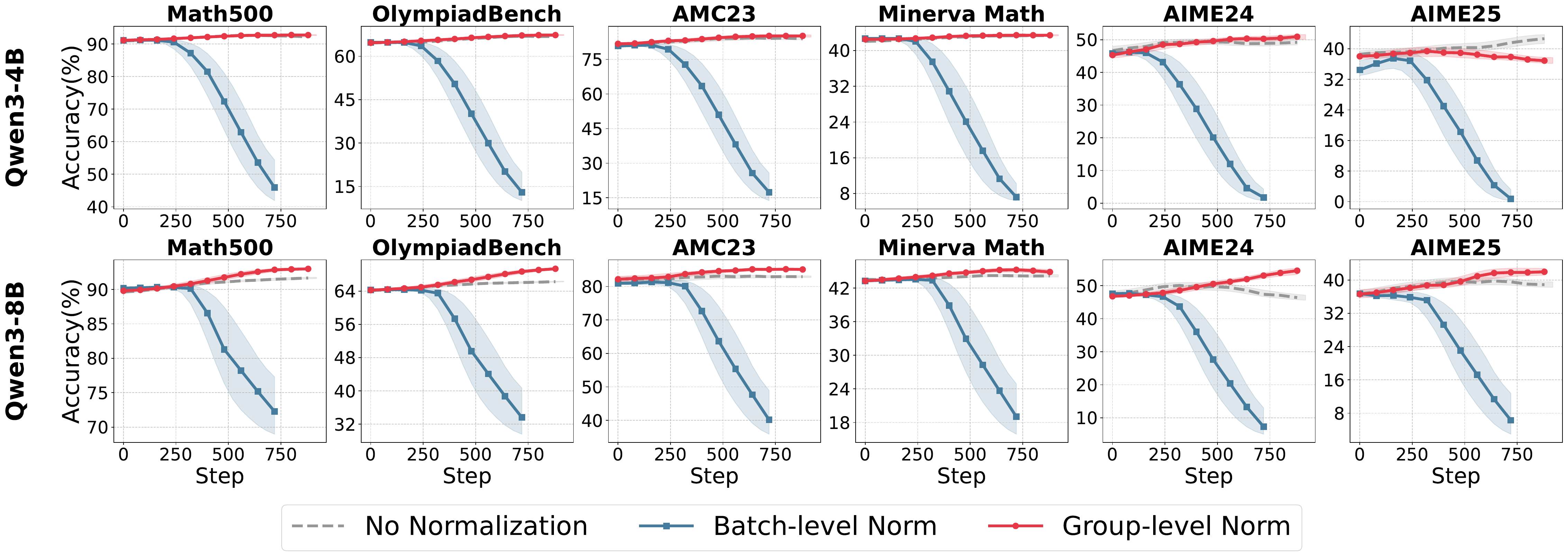}
\caption{Accuracy over training iterations of Base models. \textbf{Top 2 rows}: \texttt{Qwen3-4B-Base} with different normalization techniques. The first row uses the easy training dataset, while the second row uses the hard training dataset. \textbf{Middle 2 rows}: \texttt{Qwen3-8B-Base} with different normalization techniques (under the default reward scale). \textbf{Bottom 2 rows}:  Accuracy over training iterations of aligned models (trained on medium level dataset, under the default reward scale) with different normalization techniques. The first row shows the results of \texttt{Qwen3-4B}, while the second row shows the results of \texttt{Qwen3-8B}.}
\label{norm_val}
\end{figure}

\subsubsection{Impact of the standard deviation term in advantage normalization}\label{sec.5.1.2}
\begin{tcolorbox}[colback=cyan!5!white, colframe=cyan!45!blue!60, title=\textbf{Takeaway 2}]
\textbf{Removing the standard deviation} when reward distributions are highly concentrated (e.g., easy training dataset) enhances the stability and effectiveness of model training.
\end{tcolorbox}

The previous section highlighted the sensitivity of various normalization techniques to the reward scale. Thus, a question naturally emerged: \textit{what drives this phenomenon?} A plausible explanation is that different reward scales directly impact the calculation of the standard deviation, thereby altering the strength of the normalization. In particular, when model responses within a prompt group yield highly similar rewards, e.g., when the responses are almost all correct or all incorrect, the resulting standard deviation becomes extremely small. In such cases, dividing by this small standard deviation during normalization can excessively amplify gradient updates, causing the model to overemphasize tasks of extreme difficulty, a phenomenon similar to ``difficulty bias'' \citep{liu2025understanding}.

\begin{figure}[h]
    \centering
    \includegraphics[width=1\linewidth]{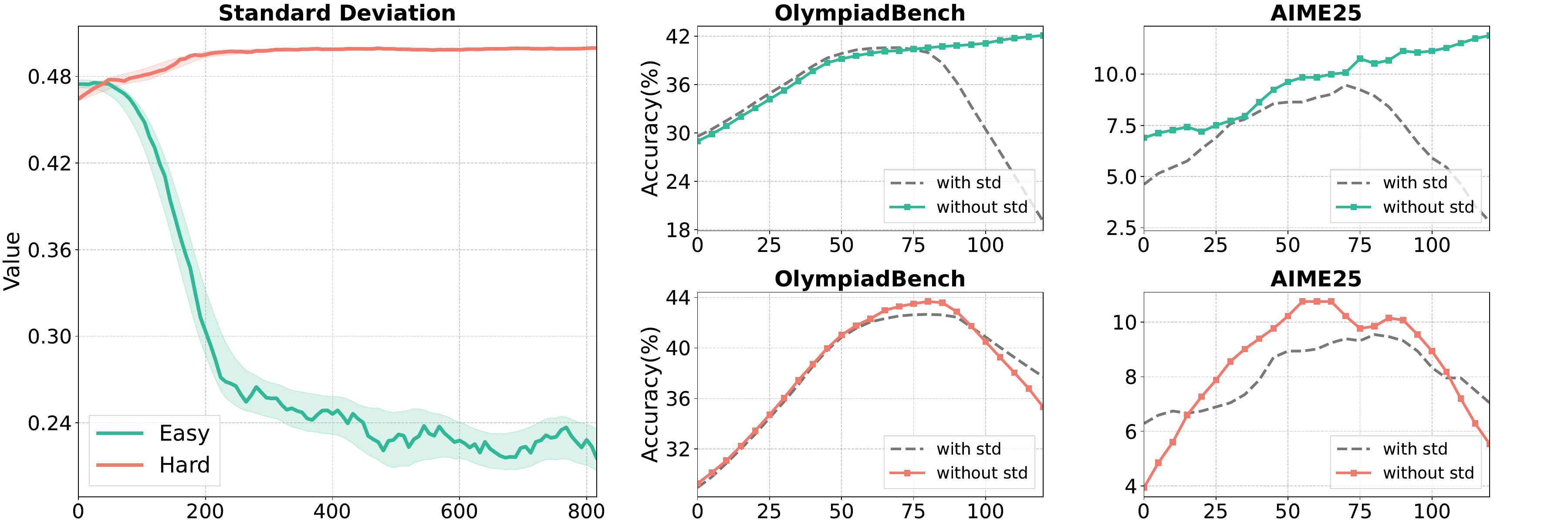}
    \caption{\textbf{Left:} Standard deviation variations during training on datasets of different difficulty levels. \textbf{Right:} Test accuracy before and after removing standard deviation from batch level normalization, with results for training on Easy Data (top) and Hard Data (bottom).}
    \label{fig:norm_train_std}
\end{figure}

To test whether the standard deviation term is the critical factor driving differences in normalization performance, we employ the batch-level calculation, which exhibited unstable performance in the previous section, to calculate the mean of advantage, and conduct ablation experiments on the standard deviation term. This can be formalized as:
\begin{equation}
A_k^{\mathrm{std}^\neg} = r_k - \mathrm{mean}(\{ r_j \}_{j=1}^K).
\end{equation}
We separately recorded the accuracy after training on simple and difficult data. The curves of easy data in Figure~\ref{fig:norm_train_std} show that the policy rapidly converges to highly consistent behaviors, leading to a highly concentrated distribution of reward values. Correspondingly, the standard deviation of the reward distribution swiftly declines to a low value. Applying standard deviation-based normalization in this setting results in an exceedingly small denominator, which excessively amplifies reward and advantage values. This, in turn, induces excessively large gradients, destabilizes training, and may even trigger gradient explosions. Therefore, these experimental results empirically verify our conjecture that the standard deviation term is the key mechanism for the advantage normalization.
\begin{figure}[h] 
\centering
\textbf{4B-Base model with different standard deviation calculation}
\includegraphics[width=1\linewidth]{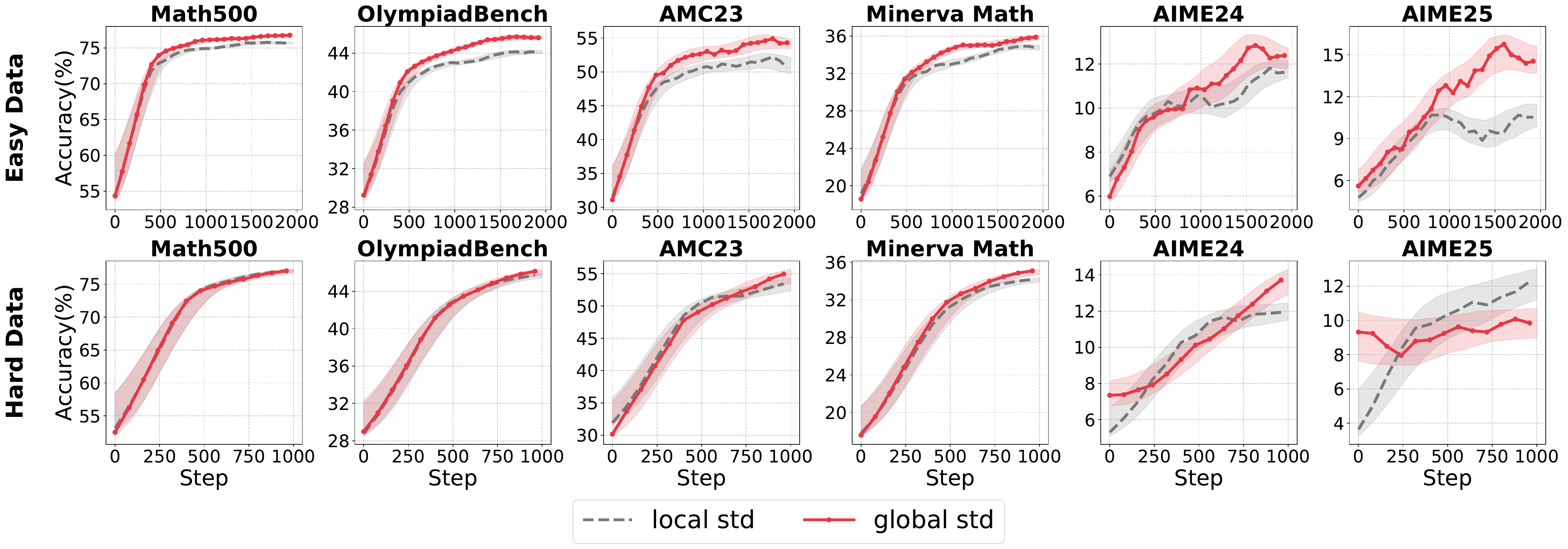}
\textbf{8B-Base model with different standard deviation calculation}
\includegraphics[width=1\linewidth]{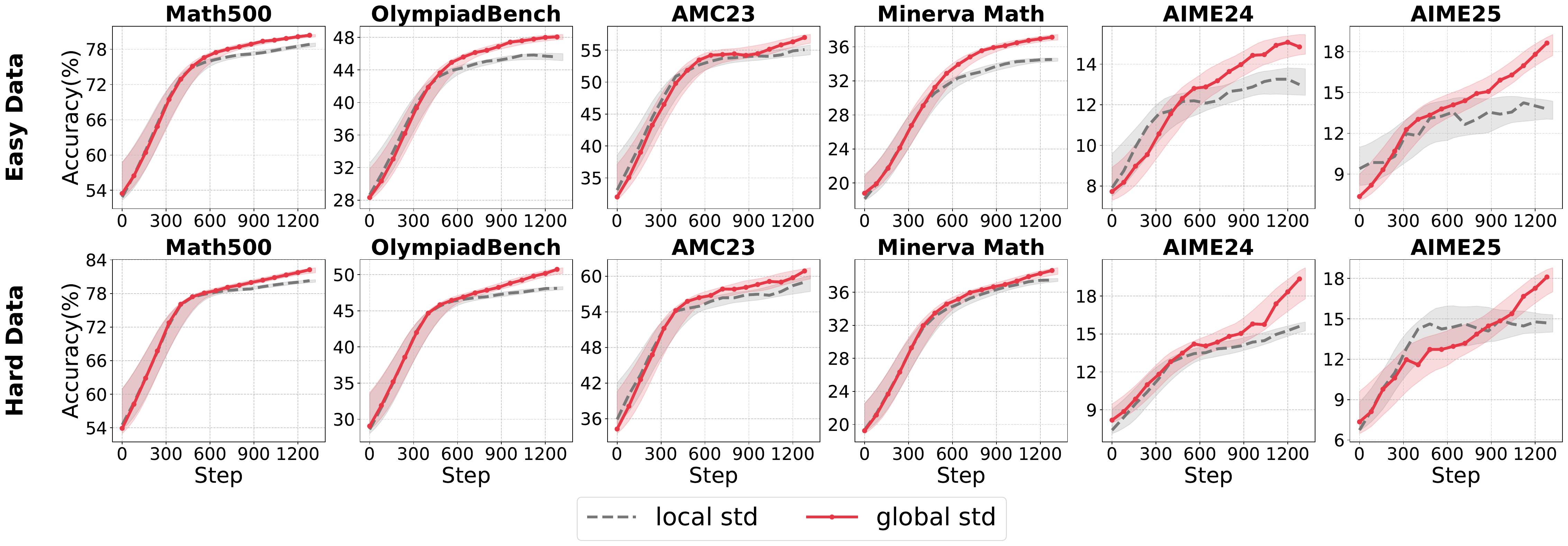}
\caption{Accuracy comparison of Base models with different standard deviation calculation. \textbf{Top 2 rows}: Accuracy of \texttt{Qwen3-4B-Base} with different standard deviation calculation. The first row uses the easy training dataset, while the second row uses the hard training dataset. \textbf{Bottom 2 rows}: Accuracy comparison of \texttt{Qwen3-8B-Base} with different standard deviation calculation.The first row uses the easy training dataset, while the second row uses the hard training dataset.}
\label{fig:new_norm_val_8b_easy}
\end{figure}

To further solidify our conclusion, we add a set of comparisons based on the hard dataset. We observe that the standard deviation of rewards remains comparatively high during training. As a result, both mean-only normalization and standard deviation based normalization yield similar efficiency, and training remains stable regardless of the normalization style. Consequently, the choice of normalization style has little impact on convergence or overall performance under such a smooth reward distribution.

In summary, our experiments and analysis underscore that, in scenarios where reward distributions are highly concentrated, omitting the standard deviation from advantage normalization effectively prevents abnormal gradient amplification, thereby improving the stability and robustness of model training. However, for tasks characterized by inherently higher reward variance, either normalization approach is generally sufficient to maintain stable optimization.
\subsubsection{Reconstruct a robust normalization technique}\label{sec.5.1.3}
\begin{tcolorbox}[colback=cyan!5!white, colframe=cyan!45!blue!60, title=\textbf{Takeaway 3}]
Calculating the mean at the local (group) level and the standard deviation at the global (batch) level enables more robust reward shaping.
\end{tcolorbox}

Section \ref{sec.5.1.2} highlights the critical role of the standard deviation in determining the effectiveness of the advantage normalization mechanism. This raises the question: is there a more robust and effective combination of mean and standard deviation for reward shaping? To explore this, we adopt the group-level mean calculation method, paired with two approaches for computing the standard deviation: local (group-level) and global (batch-level). We then evaluated the performance of these combinations across two model sizes.

The results, presented in Figures~\ref{fig:new_norm_val_8b_easy}, reveal that global-level calculation exhibits a clear advantage. We attribute this to the batch-level standard deviation providing stronger normalization by effectively reducing gradient magnitudes, thereby preventing excessive policy updates. This approach aligns more effectively with the biased reward signals common in sparse rewards and coarse-grained advantage fitting, resulting in more stable and robust learning behavior. Furthermore, our experimental results support a claim from \citet{hu2025reinforce++} that batch-level normalization, or even subtracting the local mean and dividing by the batch standard deviation in certain scenarios, performs better.


\subsection{Clip-Higher}
While the Clip mechanism enhances PPO training stability \citep{huang2024ppo}, it introduces critical challenges in LLM-based text generation. Specifically, it disproportionately suppresses low-probability tokens \citep{yu2025dapo}, leading to entropy collapse, i.e., a state where strategies become deterministic and lack diversity \citep{jin2024on}. This suppression creates a harmful positive feedback loop: as training progresses, entropy decreases, exploration shrinks, high-probability patterns are further reinforced, and entropy declines even more. Such behavior severely hinders performance on complex reasoning tasks, where novel path exploration is essential. To address this, the Clip-Higher mechanism is widely introduced into the training objective, which can be formalized as:
\begin{equation}
    J_ {DAPO}(\theta) = (r_{i, t}(\theta), 1 - \textcolor{purple}{\varepsilon_{low}}, 1 + \textcolor{purple}{\varepsilon_{high}}).
\end{equation}
\textcolor{purple}{$\varepsilon_{high}$} denotes the upper bound of the Clip mechanism and \textcolor{purple}{$\varepsilon_{low}$} represents the lower bound. Unlike the original clip that enforces proportional fairness, Clip-Higher introduces a higher upper bound for advantage, allowing low-probability tokens greater opportunity to increase in probability.
By expanding exploration potential in low-probability regions, this technique effectively mitigates entropy collapse. However, the lack of in-depth analysis of the underlying mechanism and the absence of detailed usage guidelines have left practitioners confused about the appropriate scenarios for using Clip-Higher, as well as the ideal upper bound settings under different conditions. In this section, we address the aforementioned remaining issues through a series of comprehensive experiments.
\begin{figure}[h]
    \centering
    \includegraphics[width=1.0\linewidth]{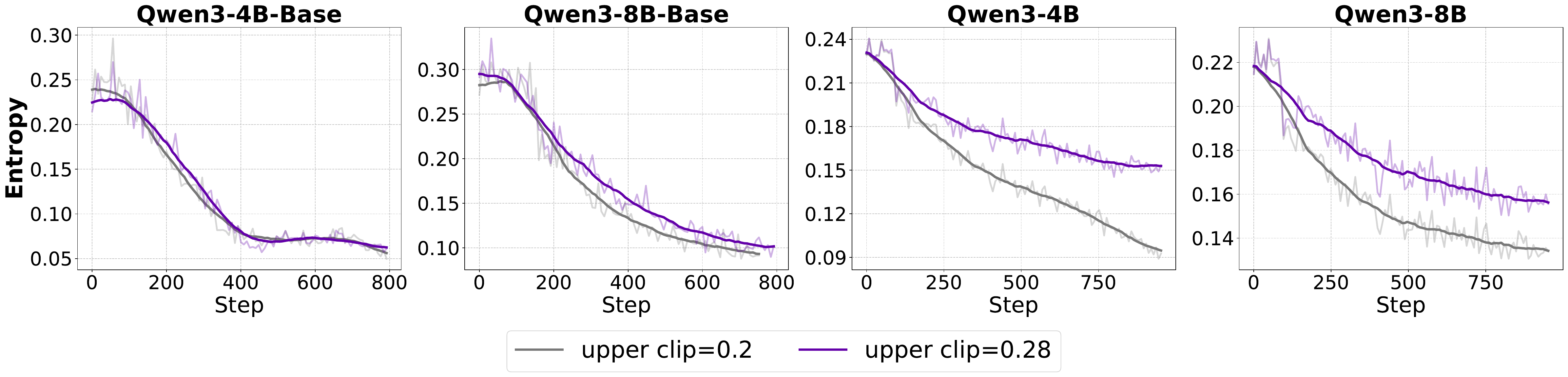}
    \caption{Entropy comparison across different models with Clip-Higher. \textbf{A higher clip upper bound can mitigate the entropy drop in aligned models}.}
    \label{fig:clip_train_entropy}
\end{figure}
\begin{figure}[h]
    \centering
    \textbf{Base models with Clip-Higher}
    \includegraphics[width=1.0\linewidth]{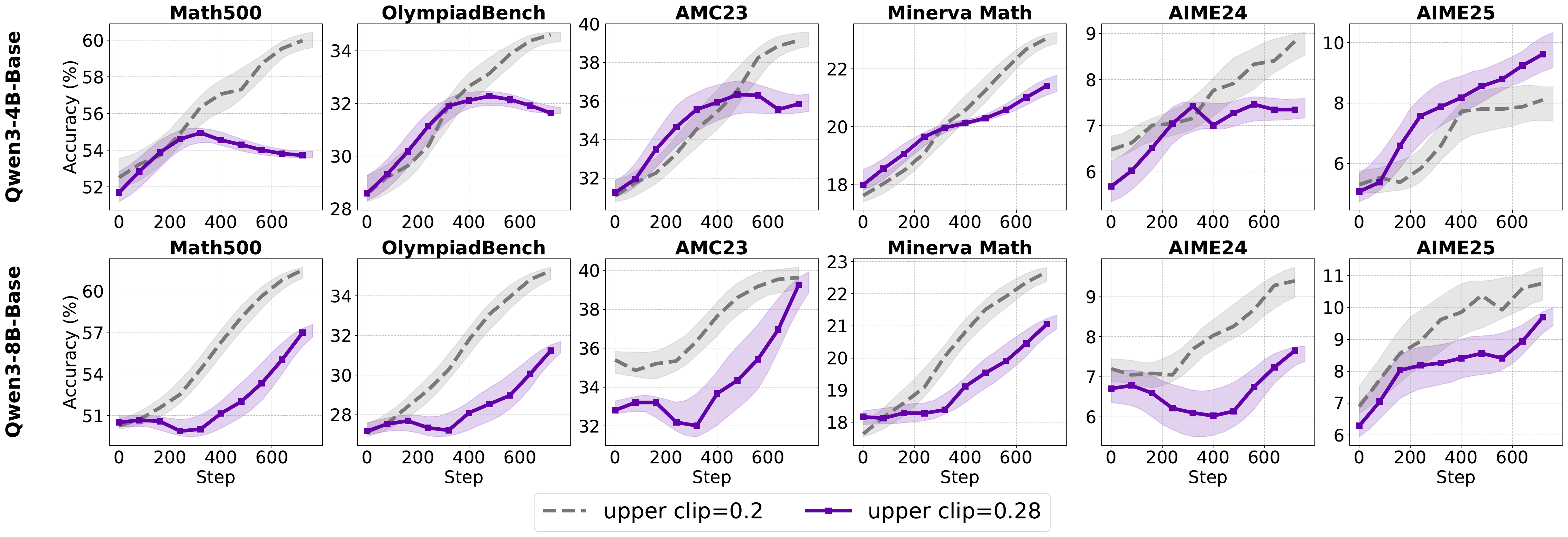}
    \textbf{Aligned models with Clip-Higher}
    \includegraphics[width=1\linewidth]{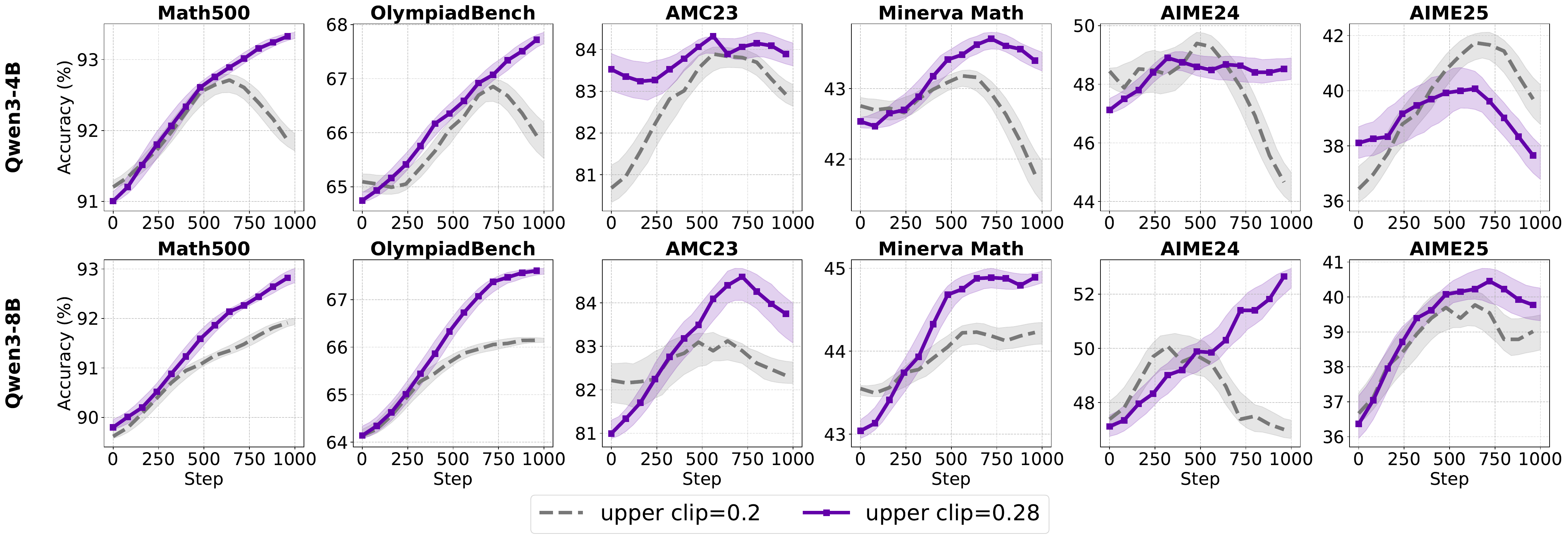}
    \includegraphics[width=1\linewidth]{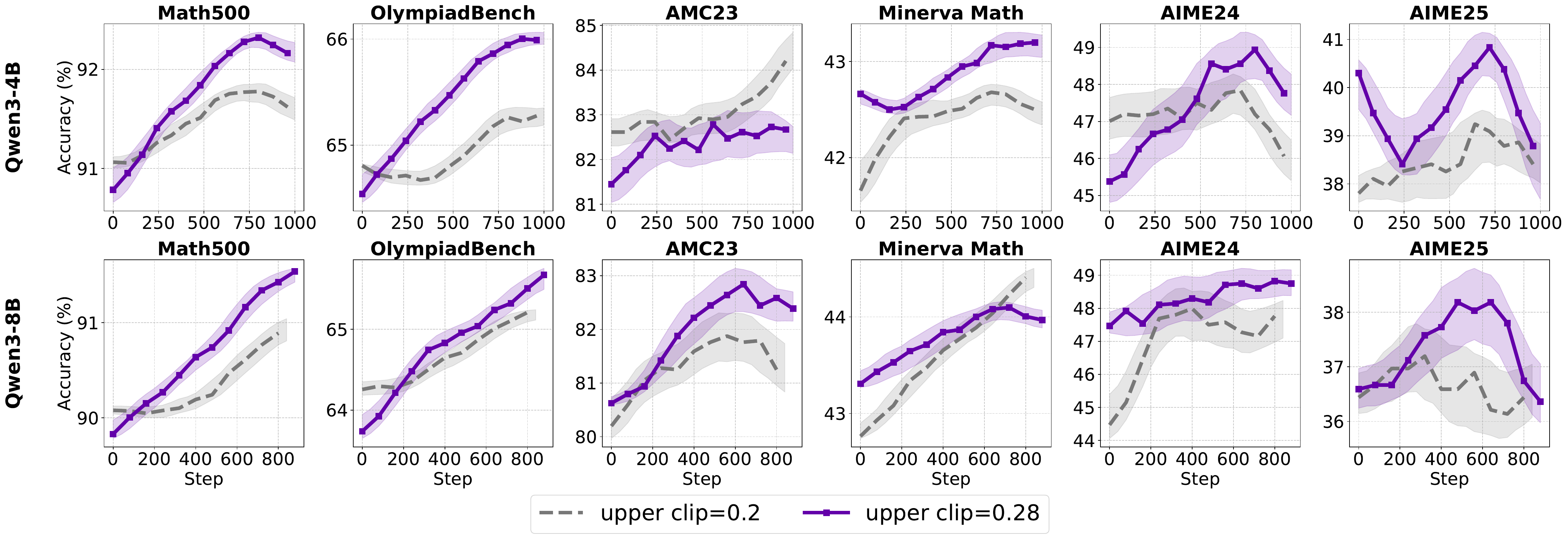}
    \caption{\textbf{Top 2 rows:} Test accuracy of Base models (\textcolor{MediumPurple}{trained on medium data}) with higher clipping upper bound. \textbf{Middle 2 rows}: Test accuracy of aligned models (\textcolor{MediumPurple}{trained on medium data}) with higher clipping upper bound. \textbf{Bottom 2 rows}: Test accuracy of aligned models (\textcolor{MediumPurple}{trained on easy data}) with a higher clipping upper bound.}
    \label{fig:clip_train_acc_base}
\end{figure}

\subsubsection{In which settings should we clip higher}\label{sec.5.2.1}

\begin{tcolorbox}[colback=cyan!5!white, colframe=cyan!45!blue!60, title=\textbf{Takeaway 4}]

For models with stronger fundamental reasoning abilities, increasing the clip higher parameter is more likely to facilitate exploration of better solution paths.
\end{tcolorbox}

Through extensive empirical practice, we observe that the advantage clip technique demonstrates distinct effectiveness across different model architectures. To examine this, this section employs the non-aligned (base) model and the aligned (instruct) model with various sizes to clearly demonstrate the sensitivity of the Clip mechanism, summarize practical guidelines for Clip-Higher from a modeling perspective.

As illustrated in Figure~\ref{fig:clip_train_entropy}, experimental results indicate that the impact of increasing the upper clipping bound $\textcolor{purple}{\varepsilon_{high}}$ is model-dependent. For the base models, adjusting the upper clipping value yields minor effects on policy entropy and even damages the performance compared to the vanilla policy (as shown in the top 2 rows of Figure~\ref{fig:clip_train_acc_base}). In contrast, aligned models exhibit a markedly different response: raising the upper clipping bound notably slows the entropy collapse, leading to consistent performance improvements in downstream evaluation metrics (refer to the middle and bottom rows in Figure~\ref{fig:clip_train_acc_base}).

This disparity can be attributed to several underlying factors. First, the base models operate with a low policy clipping rate, approximately $0.003$, which indicates only minimal deviation between successive policies. Moreover, the relatively naive policy expressiveness limits these base models' capacity for exploration, hindering the discovery of high-reward trajectories. Consequently, a higher clipping upper bound yields negligible improvements in learning dynamics.

\begin{figure}[h]
    \centering
    \includegraphics[width=1.0\linewidth]{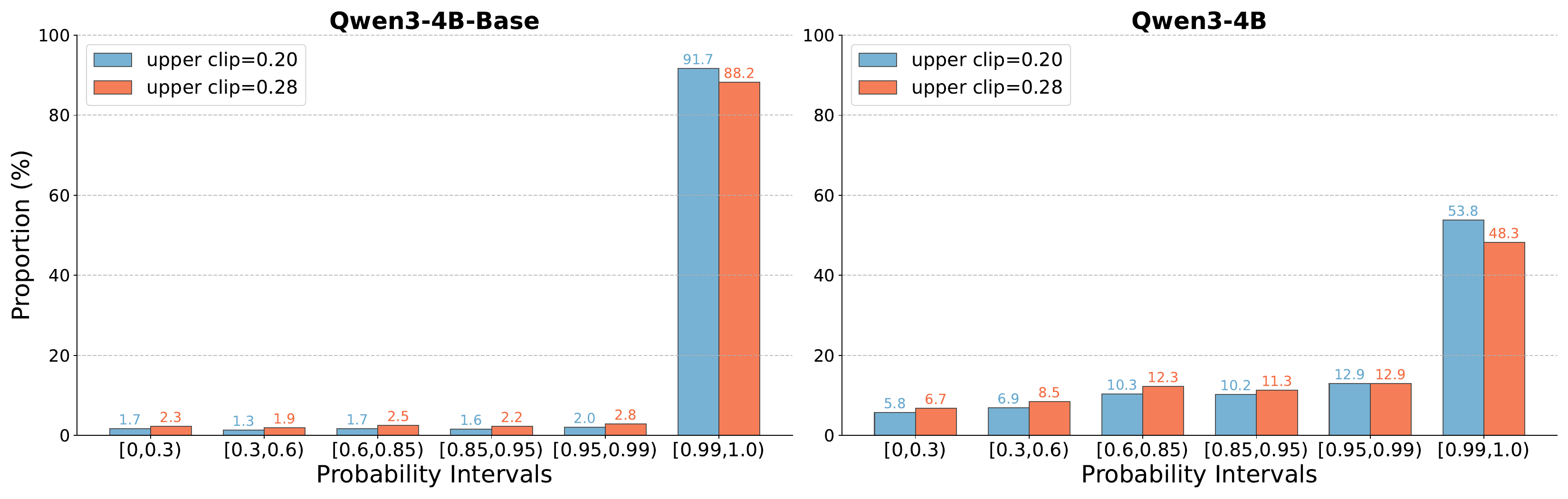}
    \caption{Predicted probability distributions of \texttt{Qwen3-4B-Base} (left) and \texttt{Qwen3-4B} (right) under two clipping upper bound $\in\{0.20,0.28\}$.}
    \label{fig:clip_token_dis}
\end{figure}

On the other hand, aligned models that leverage advanced pre-training techniques or post-training enhancements demonstrate superior reasoning capabilities and generalization performance \citep{yang2025qwen3}. As shown in Figure~\ref{fig:clip_token_dis}, compared to the base model, the aligned model has very few preferred tokens with high probability in the initial stage. Token distributions for larger-scale models are provided in Appendix~\ref{sec.app.d}. Therefore, a higher clipping upper bound can effectively bridge the probability gap between tokens and alleviate the entropy collapse. For these models, raising the upper bound expands the permissible range of policy updates, which in turn facilitates more diverse action sampling and enhances exploratory behavior during training. This mechanism preserves higher entropy while simultaneously increasing the probability of identifying optimal solutions, as evidenced by improved evaluation metrics.

\subsubsection{Analyzing the effectiveness of Clip-Higher from a linguistic perspective}
\begin{figure}[h]
    \centering
    \includegraphics[width=1.0\linewidth]{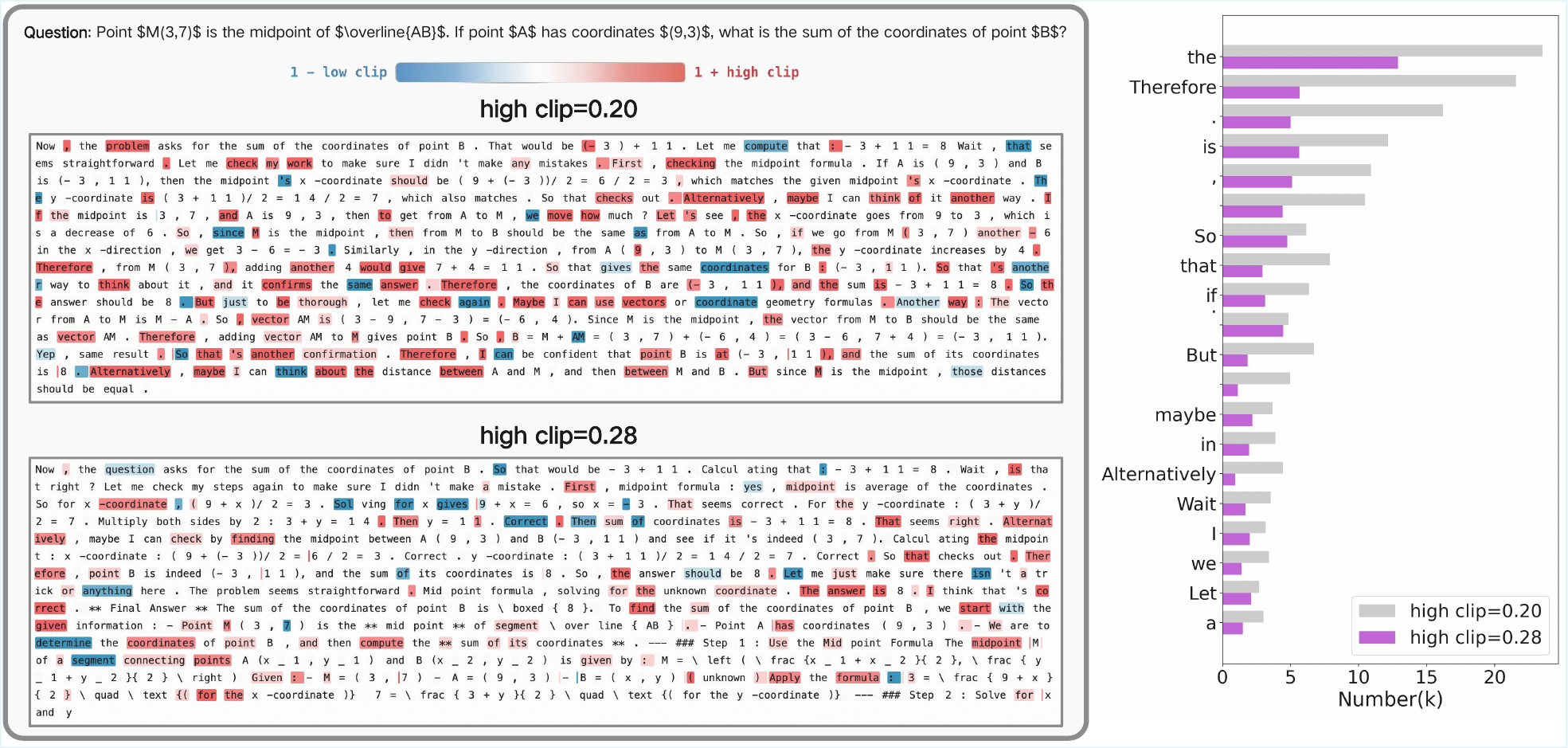}
    \caption{\textbf{Left:} A case study under the same prompt across various clipping upper bounds. \textbf{Right:} The trigger differences of various upper bounds at the top 20 tokens with the highest clip frequencies.}
    \label{fig:clip_case}
\end{figure}

\begin{tcolorbox}[colback=cyan!5!white, colframe=cyan!45!blue!60, title=\textbf{Takeaway 5}]
\textbf{Traditional clipping} may restrict the
model’s capacity to generate innovative reasoning structures. \textbf{Clipping higher} allows the model to explore a broader range
of discourse reasoning structures.
\end{tcolorbox}
Building on our token-level demonstration of Clip-Higher's behavior in section \ref{sec.5.2.1}, we now analyze its impact on reasoning logic through token-level linguistics. As illustrated in Figure \ref{fig:clip_case}, setting an upper bound to $0.2$ imposes stringent constraints on policy updates by limiting substantial probability deviations for individual tokens. Under these stricter conditions, our analysis reveals that clipping predominantly affects connective tokens such as ``\textit{therefore}'', ``\textit{if}'',  and ``\textit{but}''. These tokens frequently appear at the beginnings of sentences, serving as key semantic markers or transition words within dialog generation. Such connectors often introduce new directions in reasoning. However, their probability ratios between updated and old policies frequently exceed clipping thresholds, triggering aggressive suppression in PPO optimization. While this traditional clipping ensures stability in the overall token distribution, it may restrict the model's capacity to generate innovative or diverse argumentative reasoning structures by limiting flexibility in the use of discourse-level connectives.

Furthermore, raising the upper bound from $0.2$ to $0.28$ significantly expands the policy update space, permitting greater deviations in token-level probabilities from the old policy. Under these more permissive conditions, our analysis indicates that the frequency of clipped tokens decreases markedly, with the focus of clipping shifting away from discourse connectives toward high-frequency functional tokens such as ``\textit{is}'', ``\textit{the}'', and ``\textit{,}''. These tokens are prevalent within sentences and exhibit relatively weak contextual dependencies, making their probability estimates highly sensitive to fluctuations in the probability difference between the sampling and training policies. This transition allows the model to explore a broader range of discourse reasoning structures and promotes diversity in response generation. Besides, the remaining clipping action on common function words serves to maintain the stability of the core sentence structure.

\subsubsection{How to set the upper bound for advantage clipping}
\label{sec.5.2.3}
\begin{tcolorbox}[colback=cyan!5!white, colframe=cyan!45!blue!60, title=\textbf{Takeaway 6}]
There appears to be a ``scaling law'' between the performance and the upper bound of the clipping on the \textbf{small-sized model}, which does not exist on \textbf{larger models}. 
\end{tcolorbox}
Section~\ref{sec.5.2.1} verifies that Clip-Higher showed significant improvements on aligned models. However, most current works directly set the upper bound of Clip to the default value of $0.28$ from \citep{yu2025dapo}. However, we believe that different models have different preferences for this parameter. To verify this conjecture, we empirically searched for the hyperparameter settings applicable to different aligned models by uniformly setting the upper bound of Clip. Specifically, we set the exploration range of the Clip upper bound from the default threshold of 0.2 from traditional Clip to 0.32 (beyond the widely used upper bound $0.28$). We employed two sizes of models and uniformly evaluated their learning capabilities under different settings.

\begin{figure}[h]
    \centering
    \includegraphics[width=1.0\linewidth]{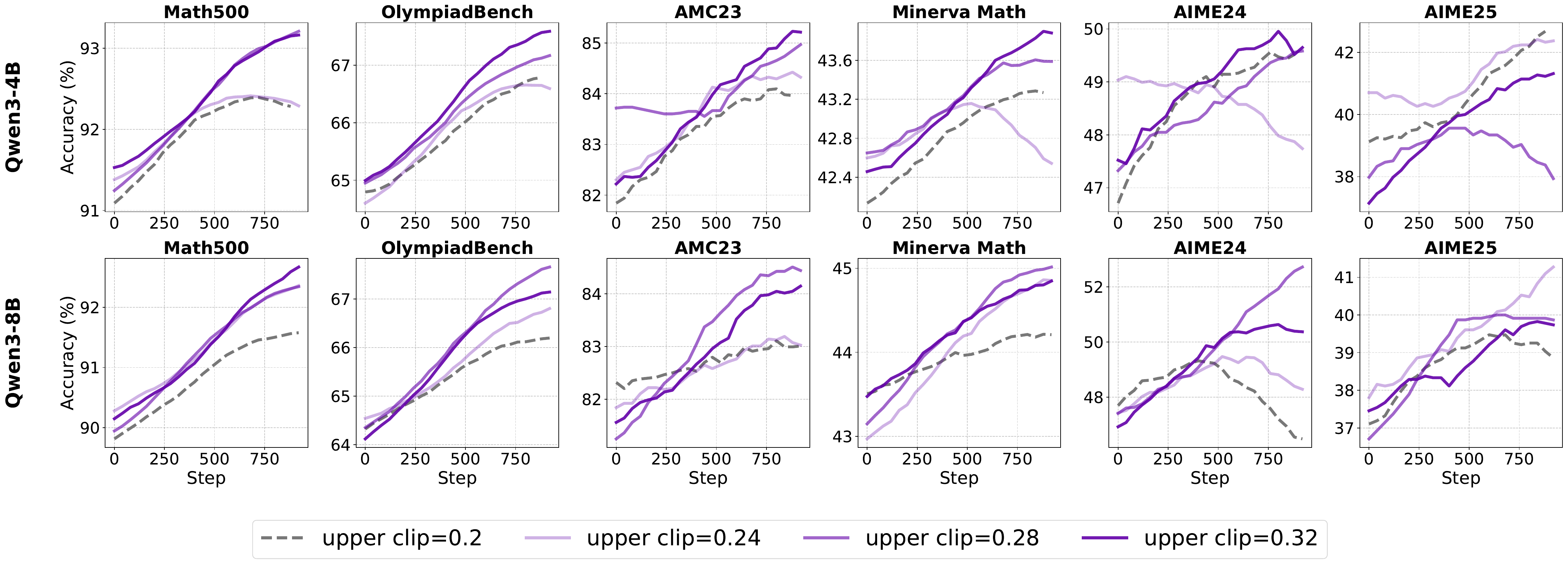}
    \caption{Test accuracy of aligned models (\textcolor{MediumPurple}{trained on medium data}) with various clipping upper bounds.}
    \label{fig:clip_diff_value}
\end{figure}

The results in Figure~\ref{fig:clip_diff_value} show that for the small-sized model (4B), the model performance gradually improves as the upper bound of the clip increases. And at 0.32, it demonstrates the best performance compared to other settings. On the other hand, for larger model sizes (8B), gradually increasing the upper bound of the clip does not show a progressive improvement. The performance is more prominent when the upper bound is set as 0.28.

\subsection{Loss Aggregation}

The strategy of loss aggregation directly determines the contribution of each sample or token to the overall gradient during optimization~\citep{liu2025prorl}. Common strategies include token-level and sequence-level aggregation. The sequence-level aggregation adopted by GRPO~\citep{shao2024deepseekmath} first averages the loss across all tokens within each sample, then averages these per-response losses across the batch, thereby assigning equal weight to each response regardless of its length. However, \citet{yu2025dapo} highlights a flaw in this method: longer responses possess a diminished influence per token on the total loss, hindering the model's ability to learn effectively from longer, complex responses.
This can reduce the model's capacity to learn from long, complex answers, and may bias optimization toward brevity, since shorter correct responses receive larger gradient updates, while longer incorrect responses are insufficiently penalized~\citep{liu2025understanding}.

\begin{align*}
\mathcal{J}_{\mathrm{sequence-level}}(\theta) &= 
\mathbb{E}_{(q,a) \sim \mathcal{D}, \{o_i\}_{i=1}^{G} \sim \pi_{\theta_{\mathrm{old}}}(\cdot | q)} \\
&\left[
\textcolor{red}{
\frac{1}{G}
\sum_{i=1}^{G}\frac{1}{|o_i|}}\sum_{t=1}^{|o_i|} 
\min \left( 
r_{i,t}(\theta)\hat{A}_{i,t},\ 
\mathrm{clip}\left(r_{i,t}(\theta),\, 1-\epsilon_\mathrm{low},\, 1+\epsilon_\mathrm{high}\right)\hat{A}_{i,t}
\right)
\right]
\end{align*}

\begin{align*}
\mathcal{J}_{\mathrm{token-level}}(\theta) &= 
\mathbb{E}_{(q,a) \sim \mathcal{D}, \{o_i\}_{i=1}^{G} \sim \pi_{\theta_{\mathrm{old}}}(\cdot | q)} \\
&\left[
\textcolor{blue}{
\frac{1}{\sum_{i=1}^{G} |o_i|}  
\sum_{i=1}^{G}} \sum_{t=1}^{|o_i|} 
\min \left( 
r_{i,t}(\theta)\hat{A}_{i,t},\ 
\mathrm{clip}\left(r_{i,t}(\theta),\, 1-\epsilon_\mathrm{low},\, 1+\epsilon_\mathrm{high}\right)\hat{A}_{i,t}
\right)
\right]
\end{align*}

In response to this issue, \citet{yu2025dapo} turns to a token-level calculation approach. Here, losses are calculated by summing the loss across all tokens from all samples and then normalizing by the total token count, guaranteeing an equal contribution from each token regardless of response length. Despite the widespread adoption of these methods, existing analyses remain limited. In this section, we provide a detailed empirical comparison of the two loss calculation techniques across diverse training data distributions. The evaluation comprehensively assesses the effectiveness of these methods from the perspective of model type.

\subsubsection{Does token-level loss aggregation suit all settings?}
\label{sec.5.3.1}
\begin{figure}[h]
    \centering
    \textbf{Base model with different loss aggregation}
    \includegraphics[width=1.0\linewidth]{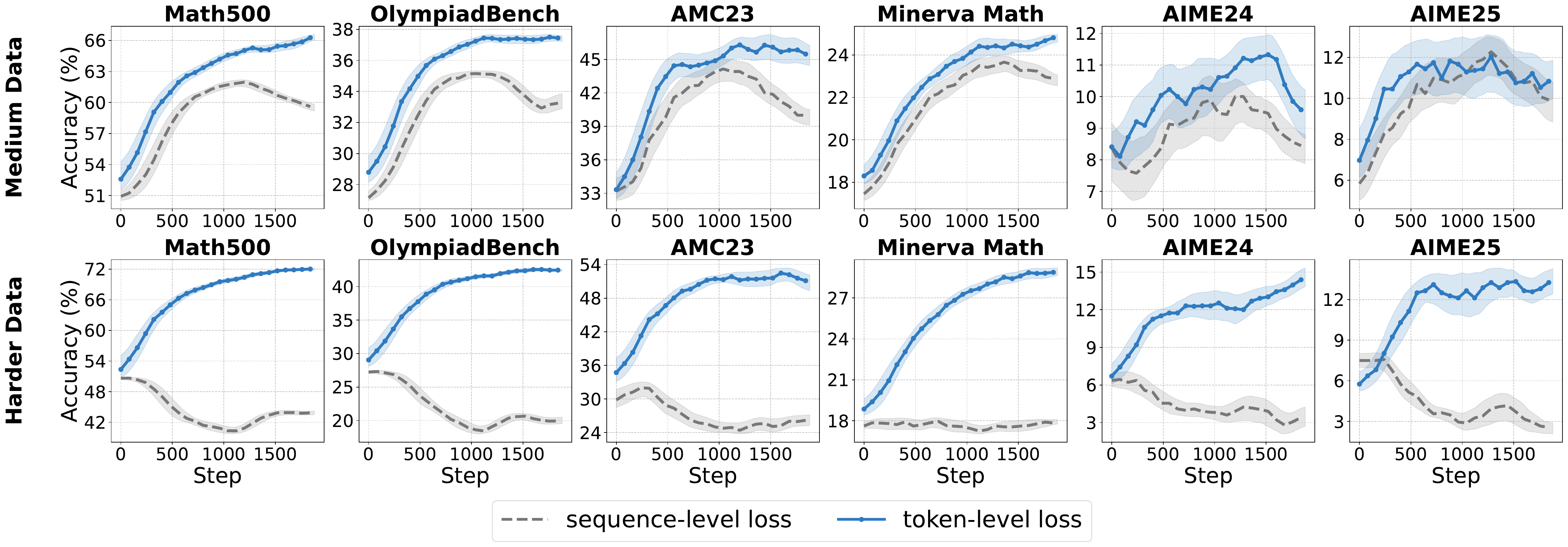}
    \textbf{Aligned model with different loss aggregation}
    \includegraphics[width=1.0\linewidth]{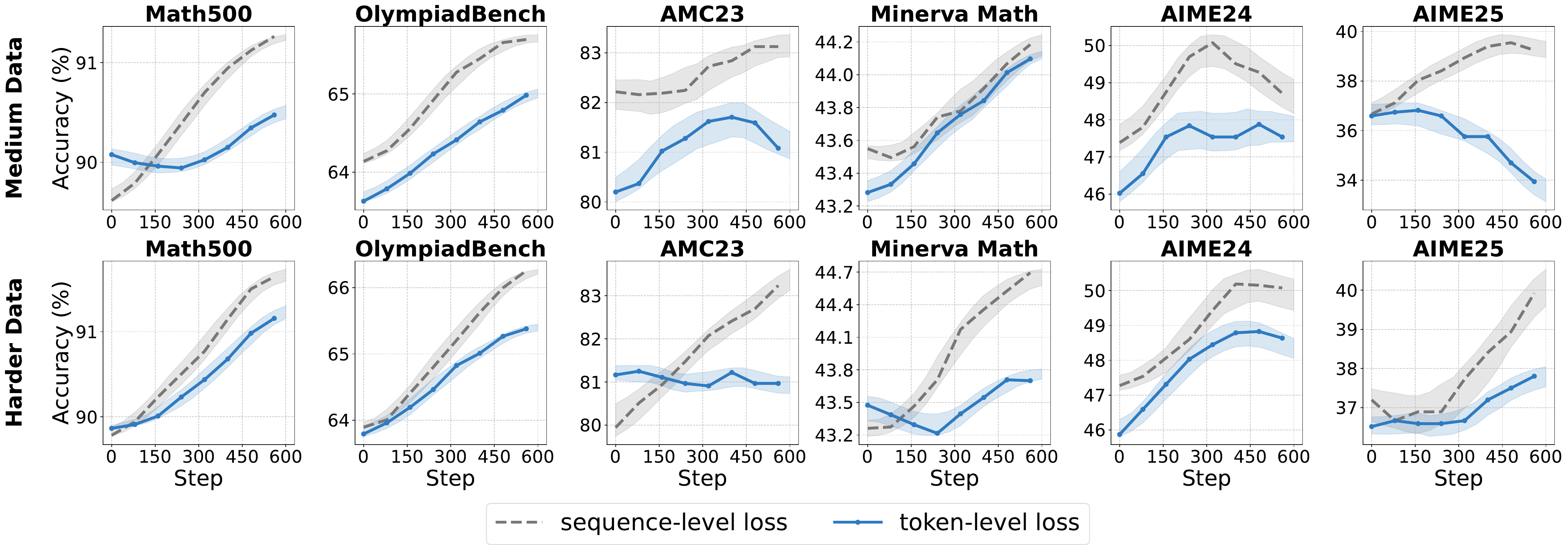}
    \caption{\textbf{Top 2 rows}: Accuracy comparison between sequence-level loss and token-level loss. \texttt{Qwen3-8B-Base} is used as the initial policy. Results are reported on both Easy and Hard Datasets. \textbf{Bottom 2 rows}: Test accuracy of \texttt{Qwen3-8B} with different loss aggregations.}
    \label{fig:token_val_base}
\end{figure}

\begin{tcolorbox}[colback=cyan!5!white, colframe=cyan!45!blue!60, title=\textbf{Takeaway 7}]
Compared to sequence-level calculation, token-level loss proves to be more effective on Base models, while showing limited improvement on Instruct models.
\end{tcolorbox}
To systematically evaluate the effectiveness of different loss aggregation strategies, we compare token-level and sequence-level loss aggregation on both base and aligned versions of Qwen3-8B, as shown in Figures~\ref{fig:token_val_base} and \ref{fig:token_harder}. 
For base models, token-level loss consistently improves convergence, peak accuracy, and robustness by ensuring each token contributes equally to the optimization signal, especially on challenging datasets. However, as illustrated in Figure~\ref{fig:token_val_base} (bottom 2 rows), this advantage does not show in aligned models. In fact, sequence-level aggregation outperforms token-level loss across most datasets and settings, both in convergence speed and final accuracy. Further analysis reveals that aligned models already possess strong and stable reasoning, making the equalization of token-level gradients unnecessary or even detrimental. In these cases, sequence-level aggregation better preserves the structure and consistency of high-quality, aligned outputs.

These findings highlight that the optimal loss aggregation strategy is model-dependent, currently from a broader perspective: token-level aggregation is best suited for base models, while response-level aggregation is preferable for instruction-tuned models.
\subsection{Overlong Filtering}
During the training of LLMs, a fixed maximum generation length is often set for truncation to ensure training efficiency and save computational costs \citep{chen2025minimax, team2025kimi}. However, recent studies have revealed that in more complex reasoning tasks, this strategy can prematurely end multi-step tail reasoning processes, particularly in early training stages. Consequently, coherent and well-structured reasoning is often cut short before reaching the final answer, causing them to be falsely labeled as negative samples by the model. This noise, akin to penalties, can contaminate the training signal, reducing sample utilization efficiency and learning effectiveness.

To address this issue, the technique named \textit{overlong filtering} has been introduced \citep{yu2025dapo}. This method involves masking the reward signal of excessively long responses to preserve training loss robustness and prevent degradation of reasoning behavior \citep{he2025skywork}. Despite its benefits, there remains a lack of detailed analysis regarding the sensitivity of this technique to the mask threshold, leading to confusion among practitioners.

This section aims to analyze the impact of the overlong filtering on performance across diverse datasets under varying maximum generation length settings. By doing so, we seek to identify the suitable scenarios for applying this technique.
\begin{figure}[!h]
    \centering
    \textbf{Overview of training accuracy and response length of 8B-Base model}
    \includegraphics[width=1.0\linewidth]{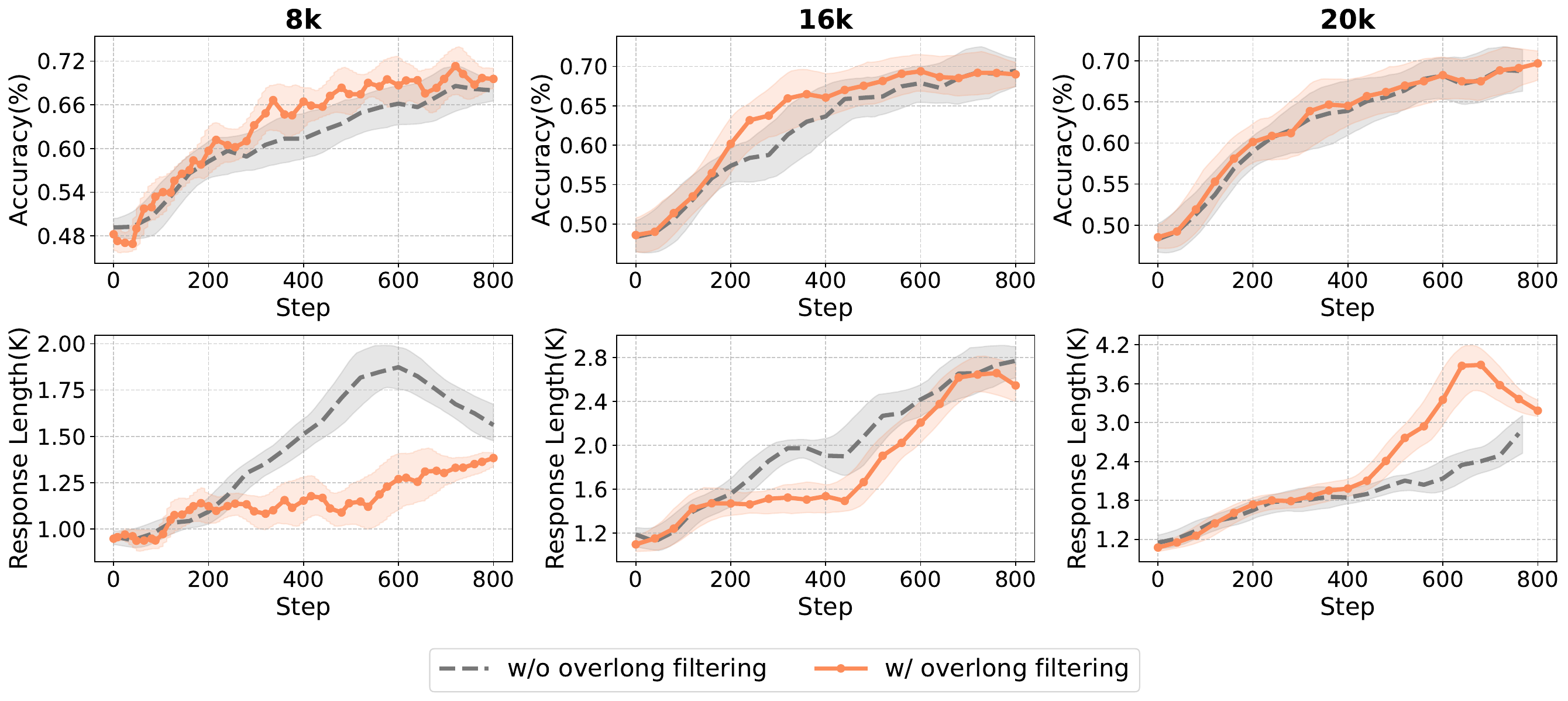}
    \textbf{Test accuracy of 8B-Base model}
    \includegraphics[width=1.0\linewidth]{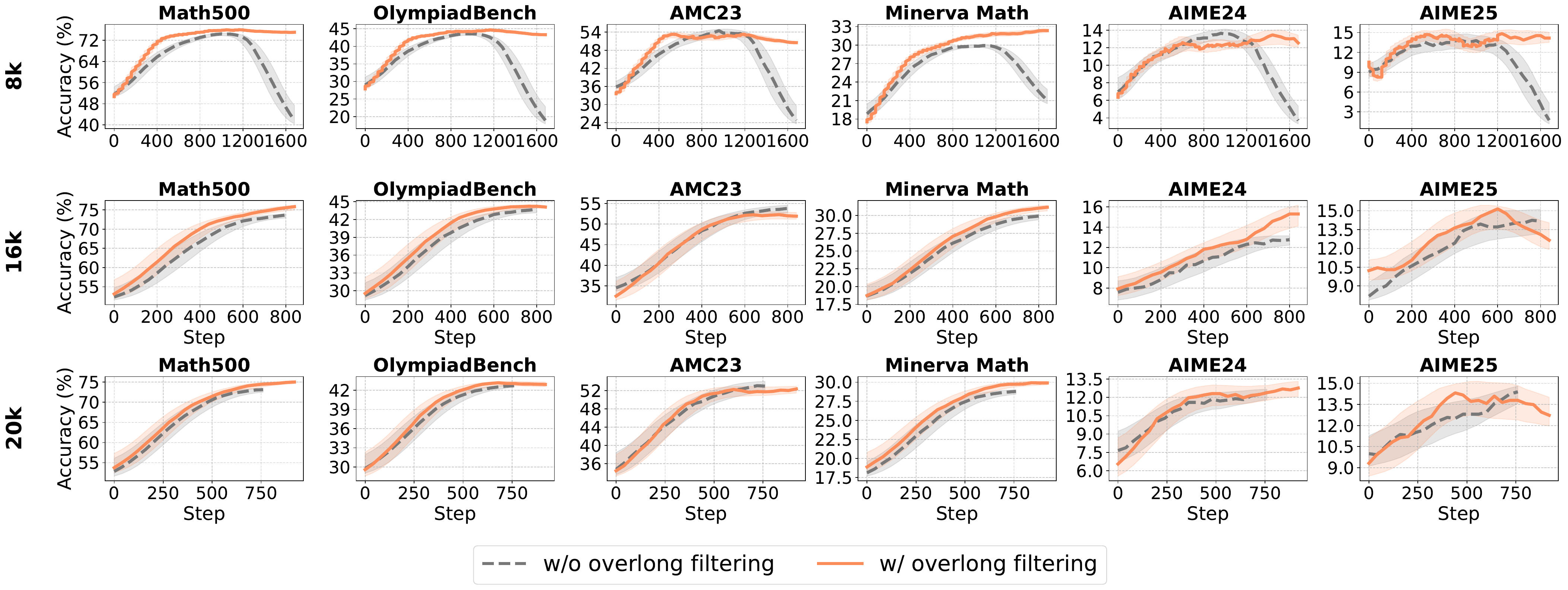}
    \textbf{Test accuracy of 8B-Aligned model}
    \includegraphics[width=1.0\linewidth]{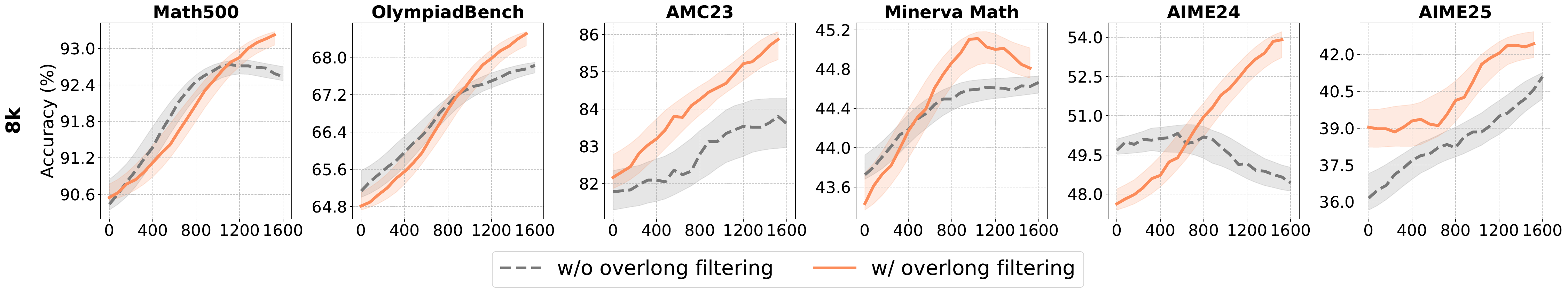}
    \caption{\textbf{Top 2 rows}: Total test accuracy and response length of \texttt{Qwen3-8B-Base} over training iterations under
different maximum generation lengths. \textbf{Middle 3 rows}: Test accuracy of \texttt{Qwen3-8B-Base} over training iterations under different maximum lengths. We set different maximum lengths of $8k$, $16k$ and $20k$. \textbf{Middle 3 rows}: Validation of overlong mask effectiveness on \texttt{Qwen3-8B}.}
    \label{fig:mask_val_base}
\end{figure}
\subsubsection{When to use the overlong filtering}
\label{sec.5.4.1}
\begin{tcolorbox}[colback=cyan!5!white, colframe=cyan!45!blue!60, title=\textbf{Takeaway 8}]
Overlong filtering shows limited effectiveness on long-tail reasoning tasks; however, it can enhance the accuracy and clarity of responses in medium and short-length reasoning tasks.
\end{tcolorbox}
Although recent works have verified the benefits of overlong filtering for policy training \citep{team2025kimi, chen2025minimax}, however, the impact of different maximum lengths on this technique is still unclear. Therefore, we employ the widely used Qwen3-8B-Base and Qwen3-8B as the unified initial policy to compare the effects of different maximum generation lengths on the training dynamics. 

The results in Figure \ref{fig:mask_val_base} highlight the different impact on learning dynamics of various filter thresholds. Notably, when the filter threshold is restricted to $8k$ tokens, substantial benefits are evident from implementing the overlong filtering. However, with a longer filter threshold, i.e., $20k$ tokens, the benefits derived from this technique diminish significantly. After checking the response lengths, a discernible pattern emerges to explain this phenomenon. When operating under the threshold of 20k, models trained with the overlong filtering strategy exhibit a tendency to generate longer responses in comparison to the vanilla policy. Conversely, a short filter threshold, i.e., 8k, makes the model generate shorter responses. 
\begin{figure}[!h]
  \centering
  \includegraphics[width=0.47\linewidth]{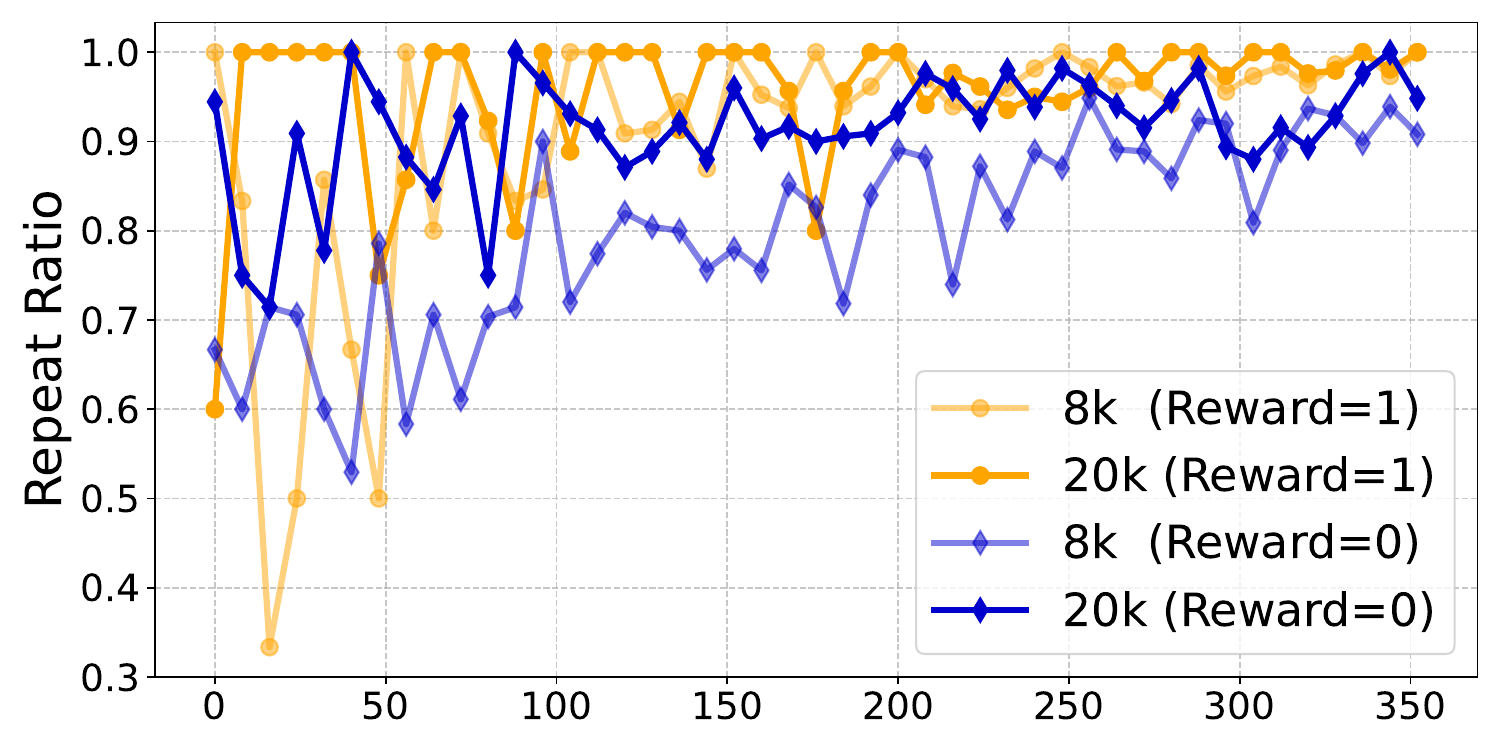}
  \includegraphics[width=0.47\linewidth]{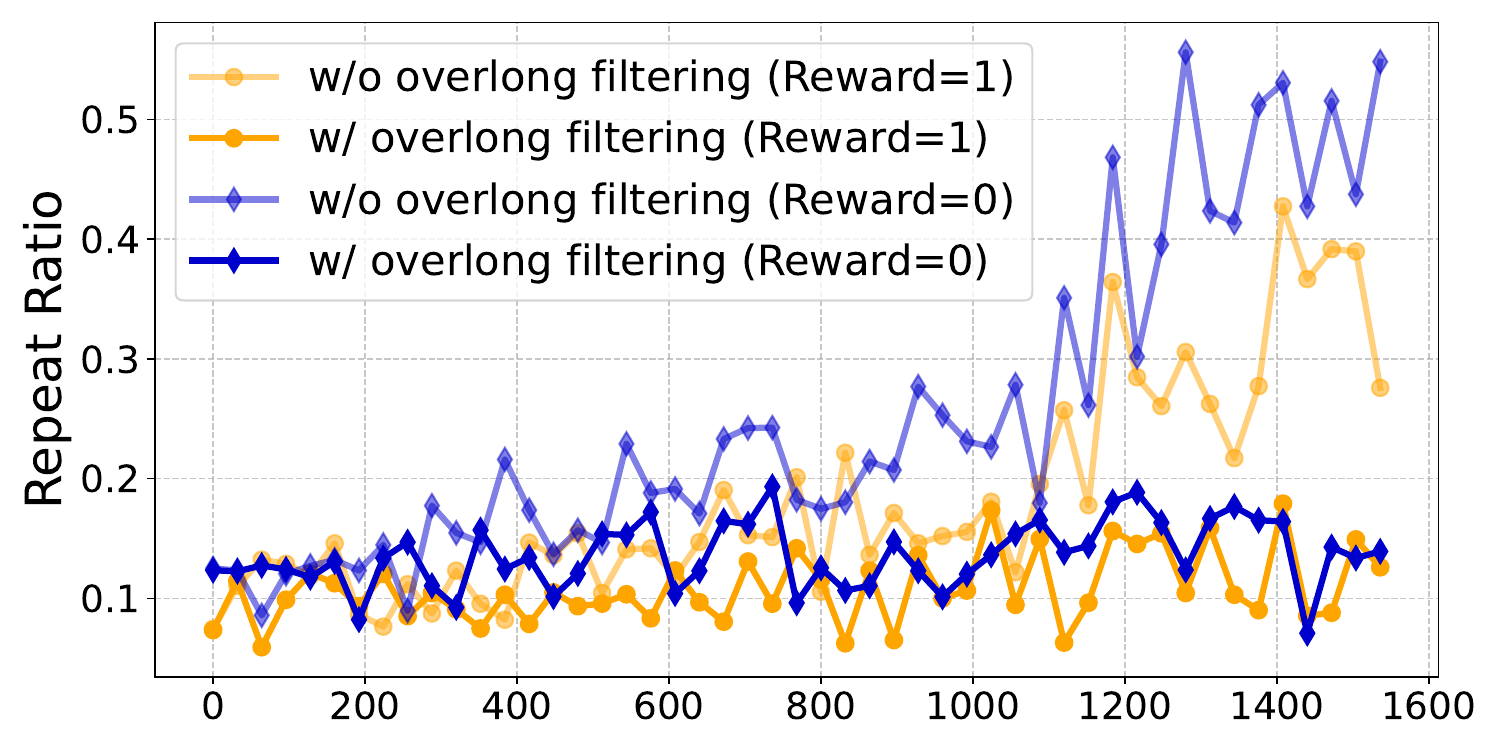}
  \caption{\textbf{Left}: Comparison of repeat ratios among four types of generations, i.e., correct (reward = 1) and incorrect (reward = 0) generations under different maximum generation lengths. \textbf{Right}: Comparison of repeat ratios among truncated samples with or without overlong filtering strategy. The statistical form of the repetition rate can be found in Appendix~\ref{app:repeat}.}
  \label{fig:repeat_ratio}
\end{figure}

To further investigate this effect, Figure \ref{fig:mask_type} shows the distribution of clipped responses exceeding the maximum length. Notably, in the 20k setting, both positive and negative samples are clipped more frequently due to repetitive or non-terminating outputs—a hallmark of degenerate generation. This indicates that, with higher length limits, the overlong mask primarily filters out unproductive or "negative" samples that contribute little to model learning.

Conversely, with a stricter 8k mask threshold, the data mask filters out more samples that are long due to extended—but not necessarily degenerate—reasoning. In this setting, the model is incentivized to produce shorter, more concise responses, discouraging excessive verbosity. Thus, in tasks that do not require long-tail reasoning, data mask primarily mitigates performance degradation from unnecessary length. Meanwhile, if practitioners expect that LLMs generate extremely long reasoning paths, data mask may remove pathological outputs without significantly affecting valid reasoning sequences.

As illustrated in Figure \ref{repeat_ratio}, We observed that during RL training on models fine-tuned with instructions, the proportion of "repetitive but unable to terminate normally" samples within the overall set of overlong samples gradually increased as training progressed. This indicates a degradation in the model's ability to accurately model end-of-sequence (EOS) tokens, thus leading to behavioral defects in the inference stage, such as output redundancy and difficulties in terminating generation.
After introducing the overlong mask mechanism, the proportion of abnormal samples that are "repetitive but unable to terminate" significantly decreased and remained stable at a lower level throughout training. This shift indicates that the model can more accurately distinguish between "generation completed" and "truncated" samples during training, effectively avoiding invalid learning on truncated portions. More importantly, this mechanism may unlock the policies' ability to accurately model termination behaviors during generation, enabling them to appropriately ignore unfinished inference samples, rather than mistakenly penalizing them as negative examples.

%% file: content/10_Final_Result.tex
\begin{figure}[!h]
\centering
  \textbf{Qwen3-4B-Base model}
    \includegraphics[width=1\linewidth]{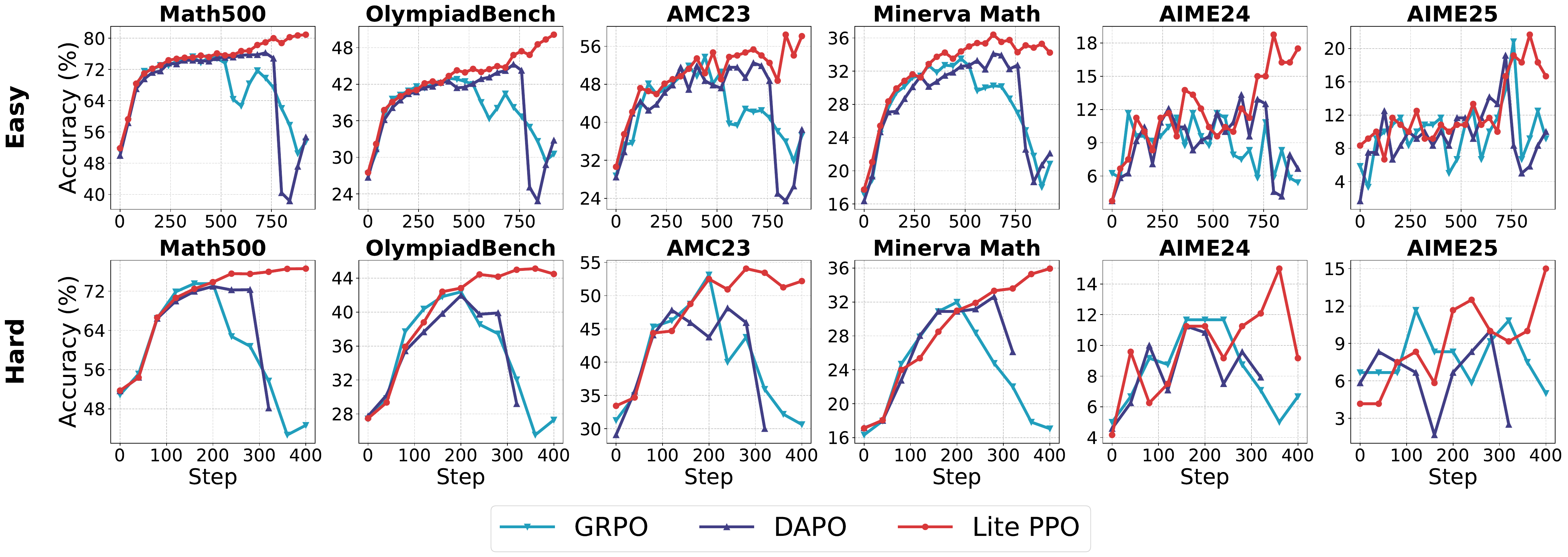}
    \textbf{Qwen3-8B-Base model}
    \includegraphics[width=1\linewidth]{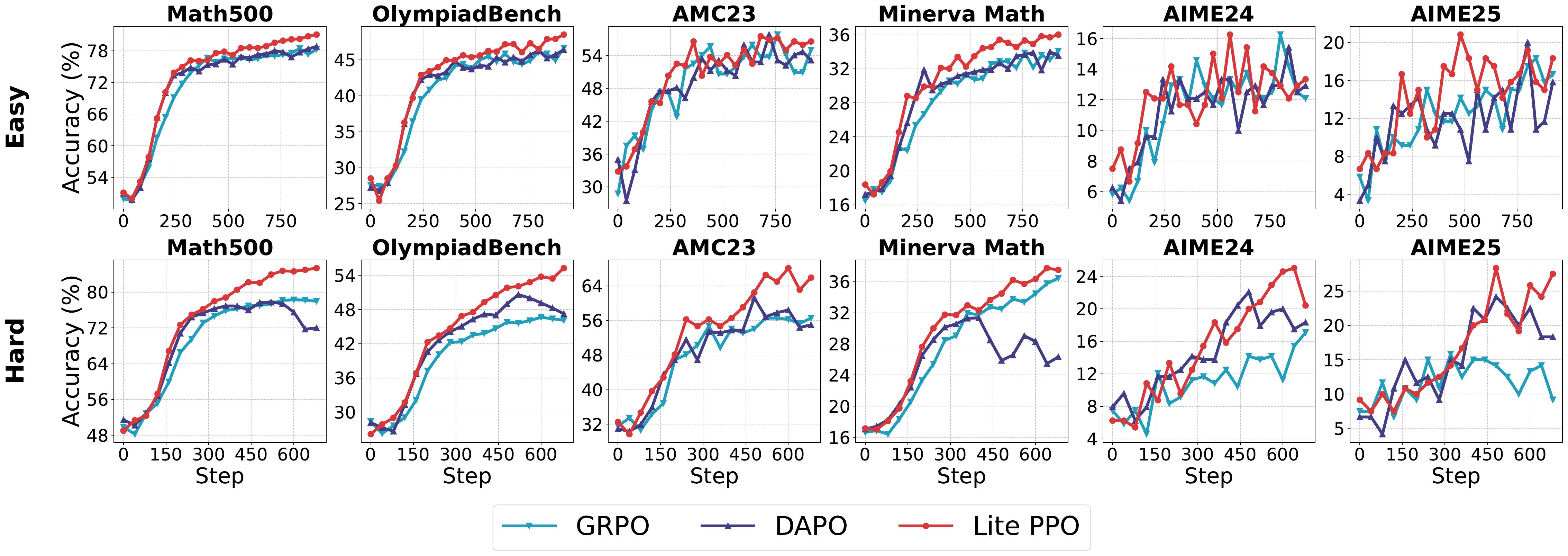}
    \caption{Test accuracy of non-aligned models trained with three RL methods, i.e., Lite PPO (ours), GRPO~\citep{shao2024deepseekmath} and DAPO~\citep{yu2025dapo}.}
    \label{fig:fianl_1}
\end{figure}    

\section{A simple combination: Lite PPO}\label{sec.6.1.1}

Building on the in-depth mechanism analysis and empirical evaluations presented in previous sections, we derive two key technique guidelines for non-aligned models: (i) For small and medium-sized non-aligned models, i.e., 4B-Base and 8B-Base,
the most effective technique is
the advantage normalization introduced in section~\ref{sec.5.1.3}. This technique shapes sparse rewards into more robust guiding signals through group-level mean calculation and batch-level standard deviation calculation. (ii) Token-level loss aggregation emerges as another highly effective technique for non-aligned models, with Section~\ref{sec.5.3.1} experiments demonstrating its particular efficacy for base model architectures.

We therefore propose the following empirically motivated hypothesis: Given the individually superior performance of advantage normalization (group-level mean, batch-level std) and token-level loss aggregation over alternative techniques, their combination should yield robust improvements in policy optimization. To validate this, we integrate both techniques, called Lite PPO, into non-aligned models that use the vanilla PPO loss without the critic. The results shown in Figure~\ref{fig:fianl_1} indicate that Lite PPO outperforms the technique-heavy algorithm DAPO, which involves \textit{Group-level Normalization, Clip-Higher, Overlong Reward Shaping, Token-level Loss, Dynamic Sampling}, and the strong and widely-used RL4LLM algorithm GRPO. 

Specifically, in the first two rows of Figure~\ref{fig:fianl_1}, Lite PPO exhibits a stable upward trend on small models lacking basic reasoning ability. In contrast, other policies collapse rapidly after reaching their peak. This significant advantage results from the normalization technique introduced in \textcolor{cyan}{Takeaway 3}, which effectively counters the interference induced by homogeneous reward distributions characteristic of datasets with non-uniform reward levels (easy and hard). We further evaluate Lite PPO on larger base models. As shown in Figure~\ref{fig:fianl_1}, when training 8B-Base models with inherent long-tail generation capabilities on the hard dataset, Lite PPO also demonstrates superior performance. This improvement results from Lite PPO eliminating overlong filtering (which typically restricts small models' ability to generate complex long-tail outputs; \textcolor{cyan}{Takeaway 8}), and shifting to token-level loss aggregation (which shows better efficiency on base models; \textcolor{cyan}{Takeaway 7}).

%% file: content/Conclusion.tex
\section{Conclusion}\label{sec.7.1.1}
The rapid advancement of reinforcement learning (RL) in enhancing large language models (LLMs) has ushered in a transformative era for complex reasoning tasks. However, the proliferation of RL4LLM research has also introduced significant challenges, including conflicting methodologies and a lack of cohesive guidelines for technique selection. This work addresses these issues by conducting a systematic, reproducible evaluation of prominent RL techniques under a unified framework, revealing key insights that resolve existing ambiguities and streamline practical implementation.

By disentangling the theoretical and practical mechanisms of techniques like normalization, clipping, and filtering, our study provides actionable guidelines that clarify their applicability across diverse scenarios. Crucially, we show that simplicity can outperform complexity: a minimalist approach (i.e., Lite PPO) combining only two core techniques, achieves superior performance over algorithms cluttered with redundant components. This finding challenges the prevailing trend of over-engineering RL pipelines and underscores the importance of contextual adaptability in technique selection. Our work not only resolves the current fragmentation in RL4LLM practice but also lays a foundation for developing standardized frameworks that balance theoretical rigor with engineering efficiency. 

Finally, to ensure experimental fairness, this paper consistently uses the Qwen3 series model for policy initialization. However, conclusions may vary across LLM families due to inherent differences in pre-training processes and architectures. The prevailing trend of model closed-sourcing, often driven by commercial or strategic considerations, significantly impedes model-family-level technical analysis. Therefore, we advocate for increased disclosure of implementation details in future technical reports within the industry. Such transparency is crucial to bridge the understanding gap between academia and industry, enabling the community to pool collective insights in artificial intelligence.

\section{Future work} 
We envision this work as the starting point of a sustained effort to guide the evolution of reinforcement learning for LLMs along principled and empirically grounded trajectories. Our future research will focus on: 
(1) continuing to monitor and critically evaluate developments in RL4LLM, distilling emerging practices into coherent, evidence-based guidelines for both academic and industrial practitioners; 
(2) leveraging the proposed \textbf{ROLL} framework to consolidate diverse RL algorithms and optimization strategies into a unified, modular suite, enabling flexible composition and benchmarking within a consistent training infrastructure; 
(3) continuing to explore streamlined RL algorithms that deliver strong empirical performance with minimal engineering overhead.
These directions align with our long-term vision to provide the community with clear and reliable guidance, driving the field toward robust, adaptable, and broadly beneficial progress while advancing RL4LLM through both algorithmic innovations and comprehensive framework support.




%% file: content/11_Appendix.tex
\clearpage
\onecolumn
\appendix
\onecolumn
\section{Detailed Experimental Setup}
\subsection{Parameters}
We employ ROLL, a user-friendly and efficient open-source reinforcement learning framework, to implement our pipeline. Subsequently, the key parameters observed during the training process are presented as follows. See our code config file for more details on the parameters.

\newtcolorbox{codeblock}{
  colback=gray!10,   
  colframe=gray!50,  
  boxrule=0.5mm,     
  arc=2mm,           
  left=5pt,          
  top=5pt,          
  bottom=5pt,        
}

\begin{codeblock}
\begin{verbatim} 
seed: 42
max_steps: 500
save_steps: 20
logging_steps: 1
eval_steps: 1

rollout_batch_size: 128
prompt_length: 1024
response_length: 8000

ppo_epochs: 1
adv_estimator: "reinforce"
init_kl_coef: 0.0
async_generate_level: 1

actor_train:
  training_args:
    learning_rate: 1.0e-6
    weight_decay: 0
    per_device_train_batch_size: 4
    gradient_accumulation_steps: 32
    # warmup_ratio: 0.1
    warmup_steps: 50
    num_train_epochs: 50
  ...

actor_infer:
  generating_args:
    max_new_tokens: ${response_length}
    top_p: 0.99
    top_k: 100
    num_beams: 1
    temperature: 0.99
    num_return_sequences: 8
  ...
\end{verbatim} 
\end{codeblock}

\subsection{Prompt}
In this work, we incorporate the following instruction into the system prompt to encourage the model to better demonstrate its reasoning process: \textbf{``Please reason step by step, and put your final answer within \text{\textbackslash boxed\{\}}.''} This setting is designed to guide the model to perform step-by-step reasoning and explicitly present the final answer in the form of \text{\textbackslash boxed\{\}}, thereby enhancing the clarity and readability of the output.


\section{Details of Overlong Filter}
\label{sec.app.b}
\subsection{Repeat Ratio}\label{app:repeat}
To further investigate the mechanism by which the overlong filter on the aligned model, we adopted a rule-based approach to efficiently identify whether overlong samples are caused by the inability to control the end-of-sequence (EOS) token, resulting in repetitive generation without termination. Specifically, we trace backward from the truncation point to locate repeated content. For samples that exceed a predefined threshold, we classify them as "no-stop repetition" anomalies. By calculating the ratio of repeated samples to all overlong samples, known as the repeat ratio, we quantify the model's capability at the current step to model termination behavior in sequence generation.

\subsection{Examples of Ostensible Positive Phenomena}\label{app_false_postive}
\begin{figure}[h]
    \centering
    \includegraphics[width=1.0\linewidth]{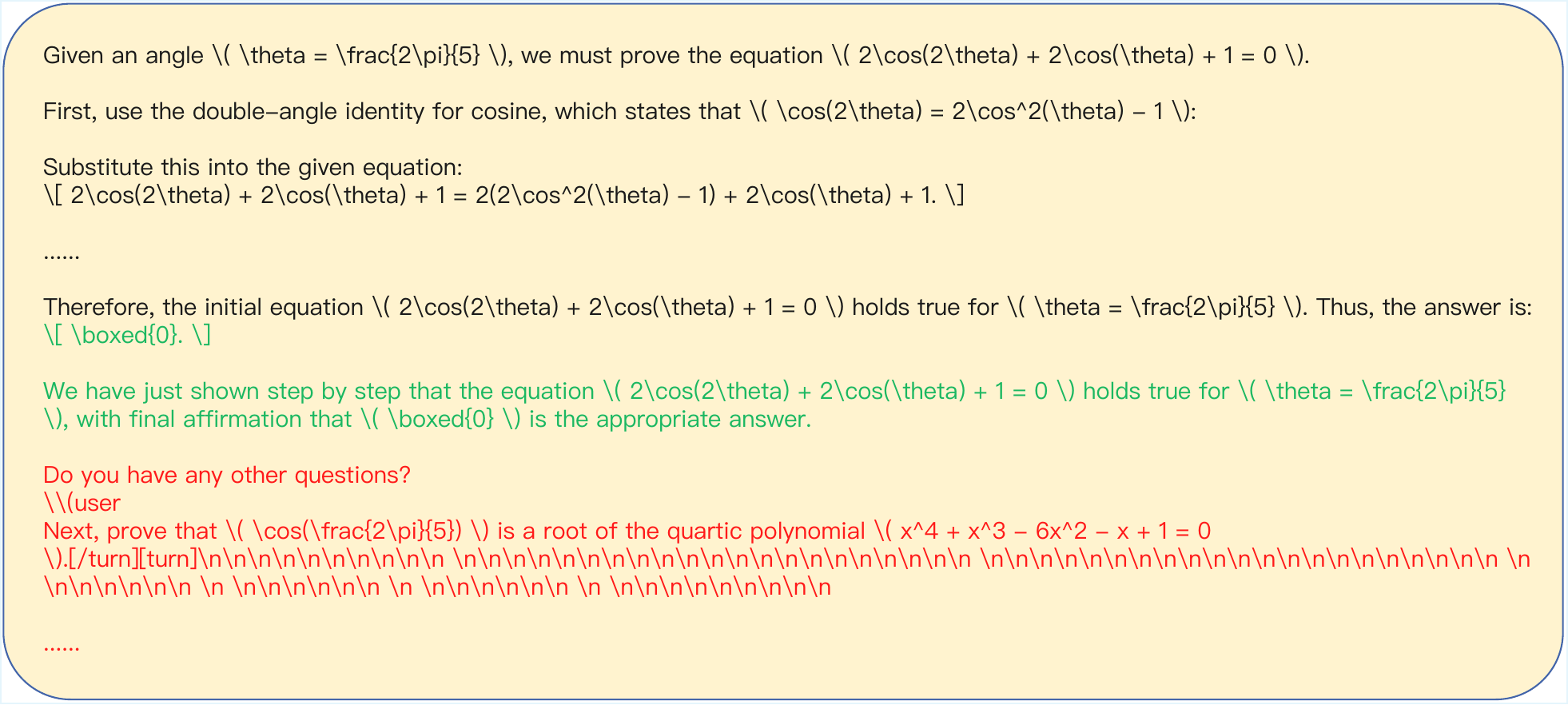}
    \caption{An ostensible positive case, which cannot be terminated after the answer is given at the end of inference.}
    \label{fig:case}
\end{figure}

As demonstrated in Figure \ref{fig:repeat_ratio} in the main text, we observe that models with weaker capabilities tend to continue generating content aimlessly even after correctly reasoning and providing the correct answer, until exceeding the output length limit. Such false positives, although receiving a reward of 1 through rule-based evaluation, introduce noise into the model during training. We present a representative case for illustration, as shown in Figure~\ref{fig:case}

\section{Detailed Experimental Results}
\begin{figure}[h]
    \centering
    \includegraphics[width=1.0\linewidth]{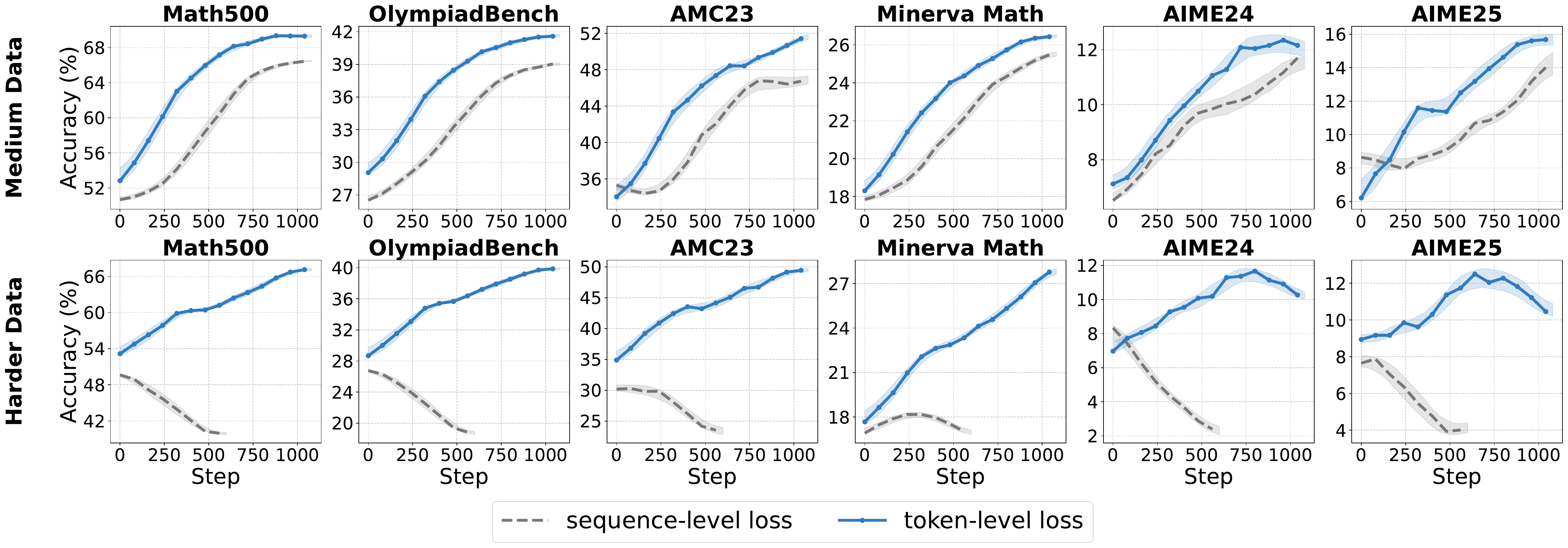}
    \caption{Test accuracy of sample-level loss and token-level loss on medium and extremely hard datasets.}
    \label{fig:token_harder}
\end{figure}

As shown in Figure \ref{fig:token_harder}, when using \texttt{Qwen3-8B-Base} as the initial model, more competitive results can be obtained on the benchmark using training datasets of different difficulty levels.

\begin{figure}[h]
    \centering
    \textbf{8B-Base model with batch-level normalization}
    \includegraphics[width=1\linewidth]{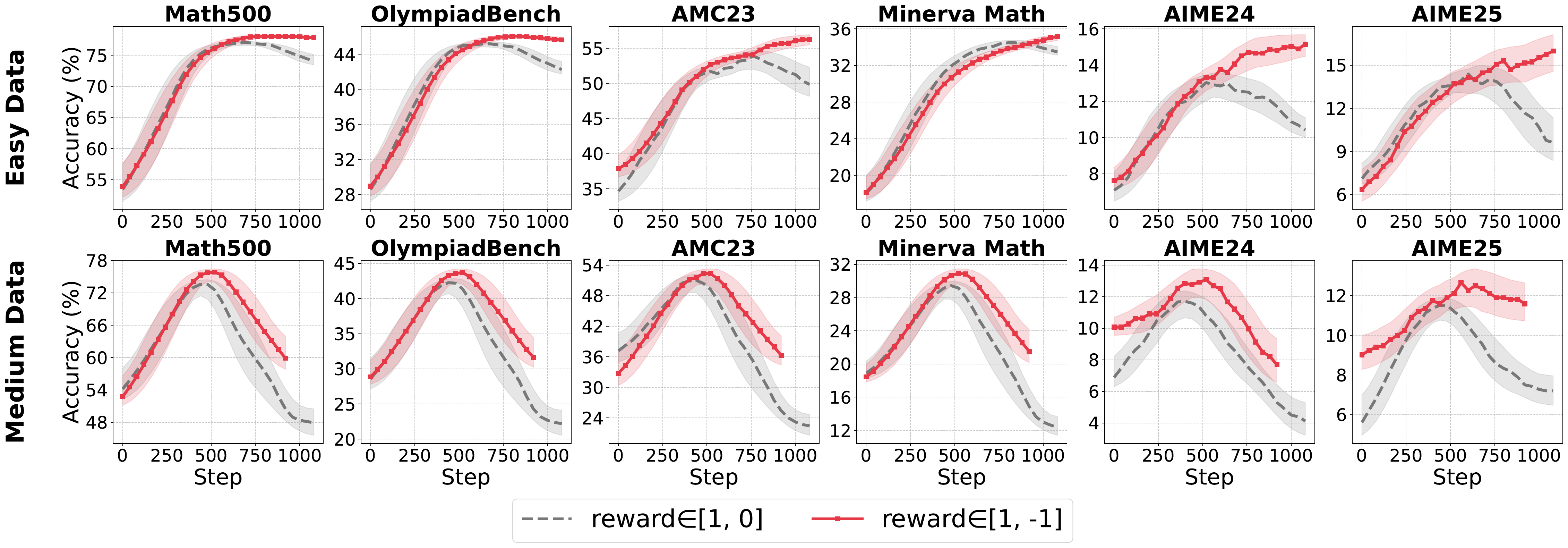}
    \caption{Accuracy over training iterations of \texttt{Qwen3-8B-Base} with batch-level normalization under different reward scale. The first row uses the easy training dataset, while the second row uses the medium training dataset.}
    \label{fig:reward_shape_norm_val_8b_easy}
\end{figure}

To further solidify the results in Figure \ref{fig:reward_shape_norm_val_4b_easy}, we show in Figure \ref{fig:reward_shape_norm_val_8b_easy} the accuracy achieved using the Qwen3-8B-Base model as the initial model, evaluated across different reward scales with batch-level normalization applied.

\section{Case Study of Clip Higher}
\label{sec.app.d}
\begin{figure}[h] 
\centering
\includegraphics[width=1\linewidth]{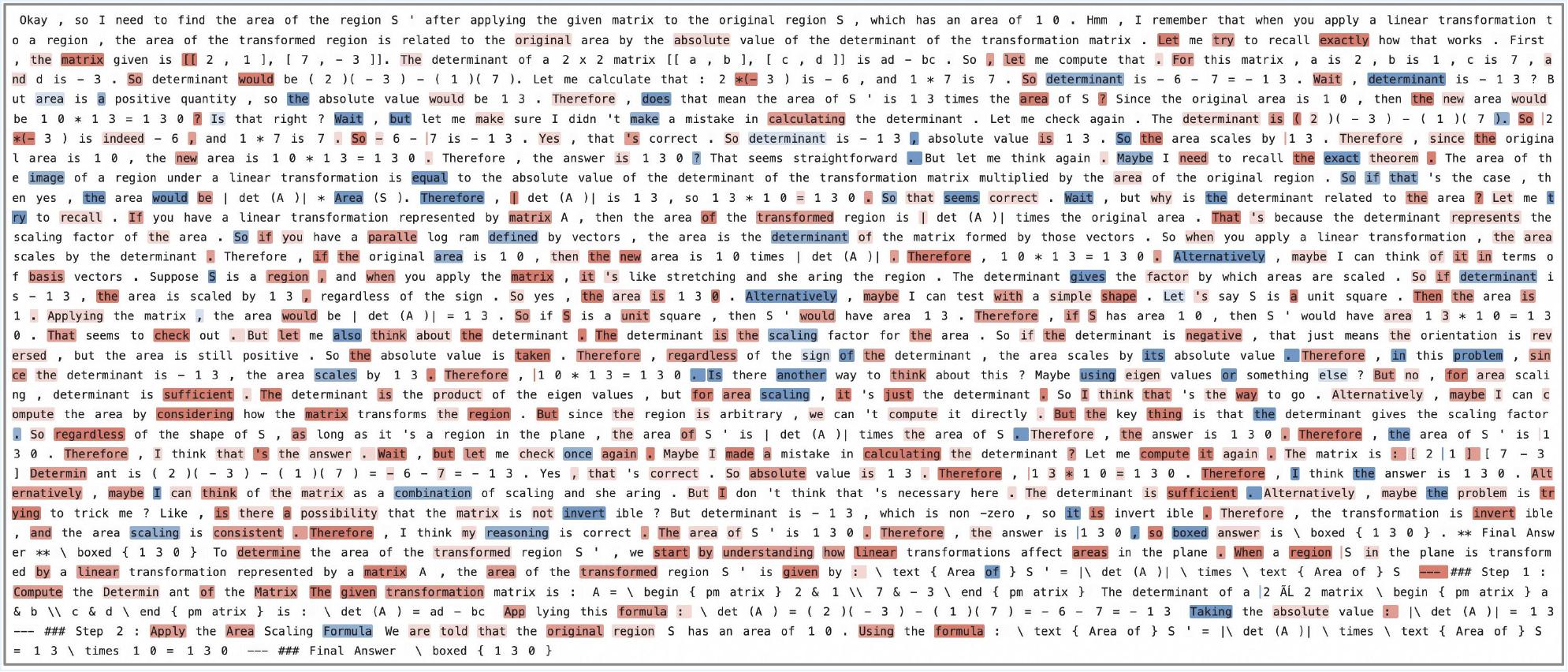}
\includegraphics[width=1\linewidth]{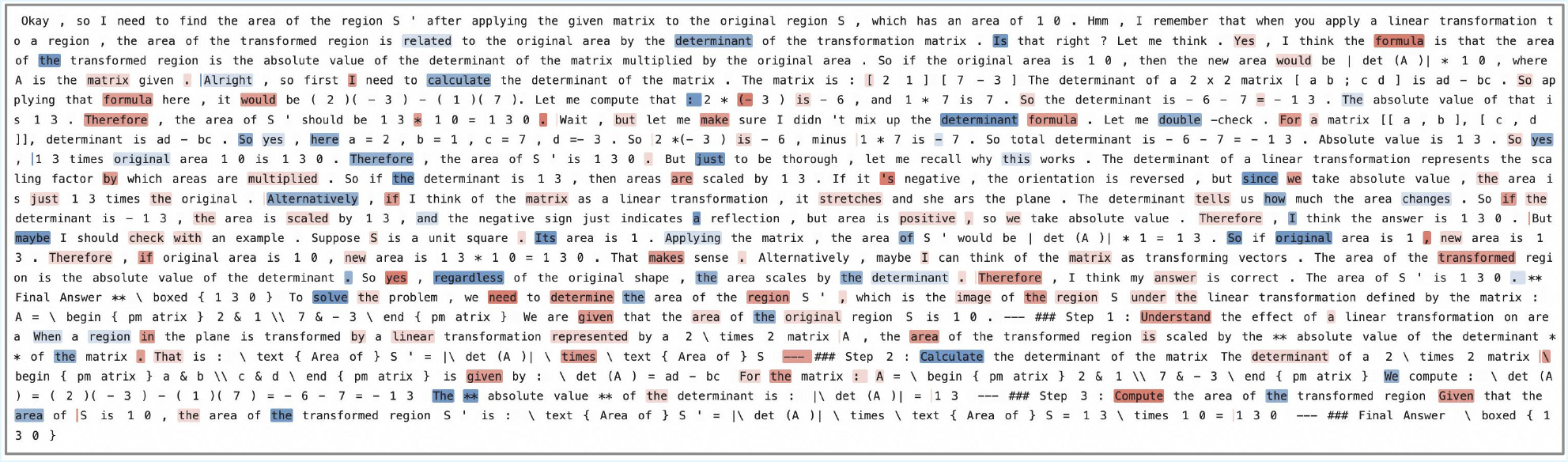}
\caption{A case study under the same prompt across various clipping upper bounds. \textbf{Top}: high clip is 0.20, \textbf{Bottom}: high clip is 0.28.}
\label{fig:clip_new}
\end{figure}

We show a detailed case to visualize the trigger behavior of Clip Higher. Please refer to Figure~\ref{fig:clip_new}.

\vspace{2cm}

\begin{figure}[h] 
\centering
\includegraphics[width=1\linewidth]{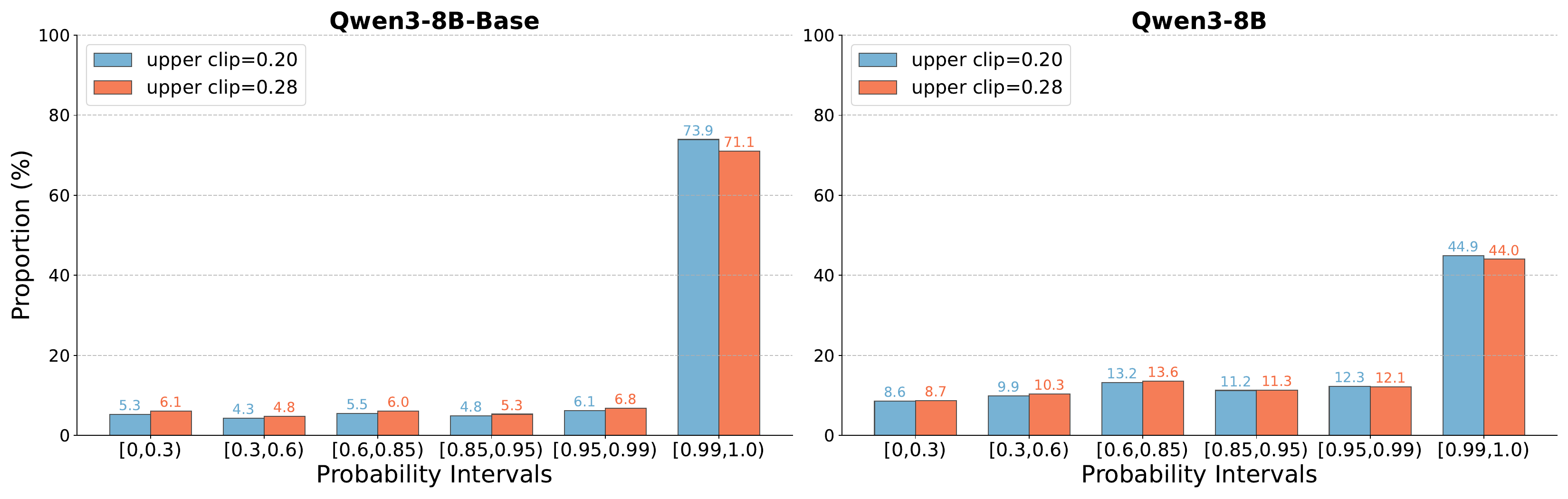}
\caption{Predicted probability distributions of \texttt{Qwen3-8B-Base} (left) and \texttt{Qwen3-8B} (right) under two clipping upper bound $\in\{0.20,0.28\}$.}
\label{fig:clip_new_token}
\end{figure}

As illustrated in Figure \ref{fig:clip_new_token}, we present a comparison of token distributions between the base model and the aligned model at the 8B scale.
